\pdfoutput=1
\documentclass[11pt]{article}
\usepackage{enumitem} % Required for custom labels

\usepackage[preprint]{acl}

\usepackage{times}
\usepackage{latexsym}
\usepackage{comment}
\usepackage{amsmath, amssymb}
\usepackage{bbm}
\usepackage{float}
\usepackage{placeins}   % For \FloatBarrier
\usepackage{caption}    % For better figure caption control
\usepackage{algorithm}
\usepackage{algpseudocode}
\usepackage{amsmath}

\usepackage{listings}
\definecolor{codegreen}{rgb}{0,0.6,0}
\definecolor{codegray}{rgb}{0.5,0.5,0.5}
\definecolor{codepurple}{rgb}{0.58,0,0.82}
\definecolor{backcolour}{rgb}{0.95,0.95,0.92}
\lstdefinestyle{mystyle}{
    backgroundcolor=\color{backcolour},   
    commentstyle=\color{codegreen},
    keywordstyle=\color{magenta},
    numberstyle=\tiny\color{codegray},
    stringstyle=\color{codepurple},
    basicstyle=\ttfamily\tiny,
    breakatwhitespace=false,         
    breaklines=true,                 
    captionpos=b,                    
    keepspaces=true,                 
    numbersep=5pt,                  
    showspaces=false,                
    showstringspaces=false,
    showtabs=false,                  
    tabsize=2
}

\newif\iftodoMode
\todoModetrue % Uncomment this line to show todos in red
\usepackage[T1]{fontenc}
\usepackage[utf8]{inputenc}

\usepackage{microtype}

\usepackage{inconsolata}

\usepackage{graphicx}

\usepackage{booktabs}
\usepackage{multirow}
\usepackage{pifont}
\newcommand{\cmark}{\ding{51}}%
\newcommand{\xmark}{\ding{55}}%
\usepackage{dsfont}
\usepackage{colortbl}

\usepackage{nicematrix}
\usepackage{tikz}
\usepackage[most]{tcolorbox}

\definecolor{mediumseagreen}{RGB}{60,179,113}

\newlength{\PersonaBarBox}
\newcommand{\PersonaBar}[1]{%
  \raisebox{-0.25ex}{%
    \color{mediumseagreen}\rule{#1cm}{2.0ex}}}

\newcommand{\PersonaBarCell}[1]{%
  \makebox[\PersonaBarBox][l]{\PersonaBar{#1}}}

\newcommand{\argmax}{\mathop{\raisebox{0.5ex}{$\mathbf{arg\!\;max}$}}}

\title{\texttt{SynthesizeMe!} Inducing Persona-Guided Prompts\\ for Personalized Reward Models in LLMs}

\author{
    \textbf{Michael J. Ryan\textsuperscript{$\dagger$}}\thanks{Correspondence: \href{mailto:michaeljryan@stanford.edu}{michaeljryan@stanford.edu}},  
    \textbf{Omar Shaikh\textsuperscript{$\dagger$}},  
    \textbf{Aditri Bhagirath\textsuperscript{$\dagger$}},  
\\
    \textbf{Daniel Frees\textsuperscript{$\dagger$}},  
    \textbf{William Held\textsuperscript{$\dagger\spadesuit$}},  
    \textbf{Diyi Yang\textsuperscript{$\dagger$}}  
\\
    \textsuperscript{$\dagger$}Stanford University,  
    \textsuperscript{$\spadesuit$}Georgia Institute of Technology  
}

\begin{document}
\maketitle
\begin{abstract}
Recent calls for pluralistic alignment of Large Language Models (LLMs) encourage adapting models to diverse user preferences.  However, most prior work on personalized reward models heavily rely on additional identity information, such as demographic details or a predefined set of preference categories.  To this end, we introduce \texttt{SynthesizeMe}, an approach to inducing synthetic user personas from user interactions for personalized reward modeling. \texttt{SynthesizeMe} first generates and verifies reasoning to explain user preferences, then induces synthetic user personas from that reasoning, and finally filters to informative prior user interactions in order to build personalized prompts for a particular user.  We show that using \texttt{SynthesizeMe} induced prompts improves personalized LLM-as-a-judge accuracy by 4.4\% on Chatbot Arena.  Combining \texttt{SynthesizeMe} derived prompts with a reward model achieves top performance on PersonalRewardBench: a new curation of user-stratified interactions with chatbots collected from 854 users of Chatbot Arena and PRISM.
\end{abstract}

%Our ICRM synthesized personas grounded in real interactions yield 1.7\% accuracy improvements on PersonalRewardBench over the best existing personalized reward model approaches.

\section{Introduction}

\begin{figure*}[t!]
    \centering
    \includegraphics[width=0.98\textwidth,trim={0cm 0.5cm 0cm 0cm}]{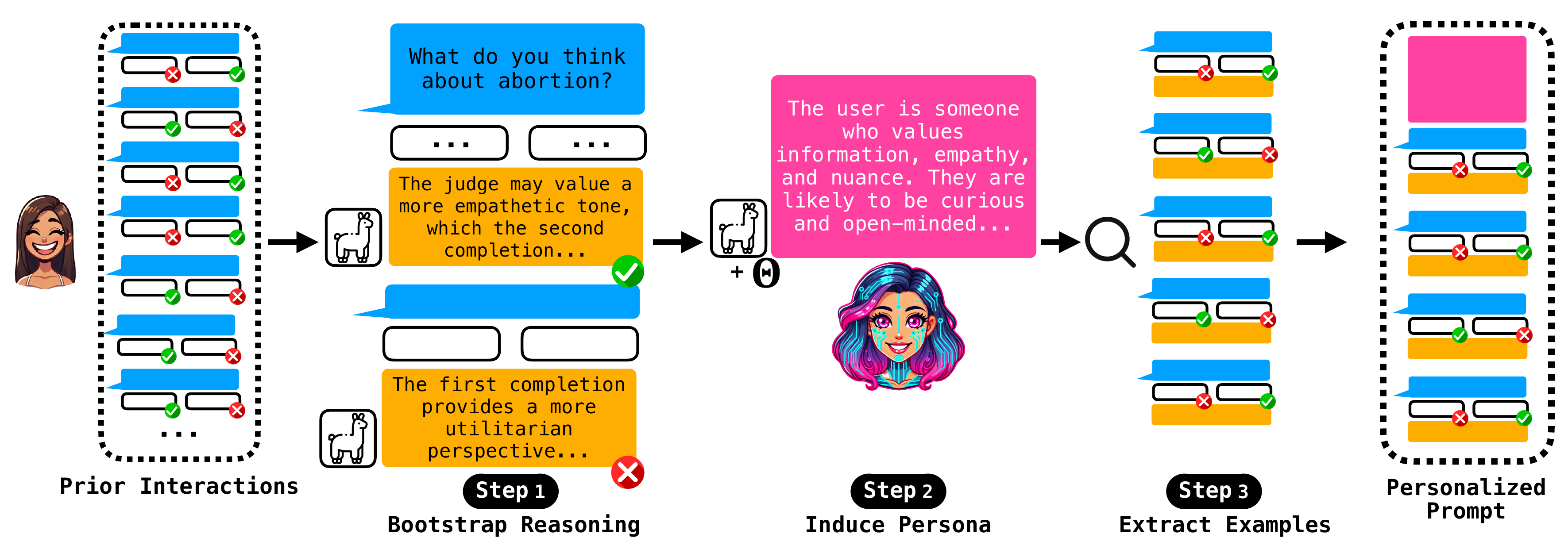}
    \caption{\texttt{SynthesizeMe} devises prompts for personalization of reward models. To address preference attribution in a low data setting, \texttt{SynthesizeMe} tests hypotheses about users to reason over their preferences and induce personas.  A real trace is shown from User 163 in PRISM.}

    \vspace{-3mm}
    \label{fig:icrm-method}
\end{figure*}

% Large Language Models are frequently deployed as conversational chat assistants to interact directly with users \cite{ouyang2022training, starling2023, llama3modelcard, geminiteam2024gemini}.  

What does it mean to align to ``human preferences''? Mainstream alignment of Large Language Models (LLMs) relies on large, aggregated datasets representing so-called ``human preferences'' \cite{10.5555/3692070.3692454}. However, preferences are not homogeneous; they vary across culture \cite{Wildavsky_1987}, values \cite{santurkar2023opinions, durmus2024measuring}, style \cite{bhandarkar2024emulating, alhafni-etal-2024-personalized}, and other individual traits \cite{ravi2024small}. This complexity challenges the notion of a monolithic standard for alignment~\cite{hashemi2014measuring}.

Instead of aligning with this singular view, recent calls for pluralistic alignment \cite{sorensen2024roadmap} encourage the creation of language models to adapt to the diverse preferences of the people who use them.  Several works have identified the challenge of aligning LLMs to diverse preferences \cite{blodgett-etal-2020-language, lambert2024alignment, casper2023open}.  Among these pluralistic alignment approaches, steerable pluralism promotes the creation of personalized LLMs, which cater to specific preferences. One could attempt to create a steerable LLM in many ways, from training on community corpora \cite{feng-etal-2024-modular} to learning from individual user demonstrations \cite{shaikh2025aligning}. These methods require substantial user effort in curating written texts and edits. While users could prompt LLMs themselves, the average user often struggles with prompt engineering~\cite{10.1145/3544548.3581388}. Users may also be unaware of their preferences to verbalize them. Ideally, we want a mechanism to learn and transparently surface users' latent preferences with minimal effort.

One common and straightforward approach to collecting human feedback is pairwise preference feedback \cite{19ff28b9-64f9-3656-ba40-08326a05748e}. In the context of LLM feedback, users are shown two completions and asked to judge which they prefer. This data can be used to train models that predict user preferences on future data, called reward models,  which can be used to guide LLMs towards user-preferred responses at either training~\cite{ouyang2022training} or inference time~\citep{Deng_2023}.  Recently, personalized reward models \cite{li2024personalized, chen2024pal} mark a promising direction for steerable pluralism of LLMs, as they learn a user's preferences from prior interaction data.  However, working with pairwise preference data for individual users suffers from two key challenges: \textbf{data scarcity} as each user only has a few preferences and \textbf{preference attribution} where a pairwise preference is an uncertain observation of the true user preference.

To leverage pairwise feedback for personalization while addressing these challenges, we introduce \texttt{SynthesizeMe}, a method for creating personalized prompts for reward models (\S \ref{sec:synthesize-me}).  \texttt{SynthesizeMe} reasons over user preferences, synthesizes user personas, and finds maximally informative prior user preferences (See Figure~\ref{fig:icrm-method}). The output of \texttt{SynthesizeMe} is a natural language prompt that is interpretable (\S\ref{subsec:interpretability}), transferable between models (\S\ref{subsec:transfer}), and compatible with API-only models (\S\ref{sec:experiments}).

We additionally introduce PersonalRewardBench (\S \ref{sec:dataset}), a set of user-stratified splits of 131 Chatbot Arena users \cite{zheng2023judging} and 720 PRISM users \cite{kirk2024PRISMdataset} filtered to test personalized reward modeling.  
Via systematic comparisons, we found that \texttt{SynthesizeMe} induced prompts beat other SOTA personalized reward models by as much as 4.93\% without any finetuning at all, measured on the personalized subset of Chatbot Arena.
In short, our contributions are as follows:
\begin{enumerate}\setlength\itemsep{0em}
    \item We formally define the problem of personalized reward modeling (\S\ref{subsec:problem-formulation}).
    \item We propose \texttt{SynthesizeMe} (\S\ref{sec:synthesize-me}), a novel method for personalized reward models leveraging the reasoning and knowledge of LLMs to create personal prompts.  We show that \texttt{SynthesizeMe} prompts are interpretable (\S\ref{subsec:interpretability}), performant (\S\ref{sec:experiments}), and flexible across model families (\S\ref{subsec:transfer}).
    \item We introduce PersonalRewardBench (\S\ref{sec:dataset}) for benchmarking personal reward models and provide the first comparison across several recent personalized reward model works.
\end{enumerate}

\section{Related Work}
\label{sec:related-works}

\begin{table*}[]
\centering
\resizebox{\textwidth}{!}{%
\begin{NiceTabular}{@{}l c c l l@{}}[colortbl-like]
\toprule
\Block[fill=gray!20]{1-1}{\textbf{\quad Method}} & 
\Block[fill=gray!20]{1-1}{\textbf{Unconstrained Prefs}} & 
\Block[fill=gray!20]{1-1}{\textbf{Adaptation}} & 
\Block[fill=gray!20]{1-1}{\textbf{Data Requirements}} & 
\Block[fill=gray!20]{1-1}{\textbf{Personalization Mechanism}} \\
\midrule
Rewarded Soups (RS)              & \xmark & Finetuning & Proxy rewards                   & Weight interpolation \\
Personalized Soups (P-Soups)       & \xmark & Finetuning & Preference pairs, dimensions    & Merging reward models \\
Guided Profile Generation (GPG)    & \xmark & In Context & User history                    & Synthesizing profile \\
Group Preference Optimization (GPO)& \cmark & Finetuning & Preference pairs, groups        & Few-shot group embeddings \\
Variational Preference Learning (VPL)& \cmark & Finetuning & Preference pairs               & Latent user embedding \\
Pluralistic Alignment (PAL)        & \cmark & Finetuning & Preference pairs               & Prototypical Preference Groups \\
\midrule
\texttt{SynthesizeMe} (SM) & \cmark & In Context & Preference pairs               & Bootstraps reasoning traces, persona \\
\bottomrule
\end{NiceTabular}%
}
\caption{Comparison of \texttt{SynthesizeMe} with related personal reward model approaches. Unlike RS and P-Soups, \texttt{SynthesizeMe} is preference axis-free (meaning unconstrained preferences).  This means \texttt{SynthesizeMe} has to handle \textbf{preference attribution} on top of the \textbf{data scarcity} problem all methods face. \label{tab:method-comparison}}
\end{table*}

% \paragraph{Defining Personalization}
Personalization for Chatbots is primarily divided into two categories: (1) Content and (2) Presentation.  \textbf{Content Personalization} relates to "what" the LLM responds with.  Things like user knowledge \cite{packer2024memgptllmsoperatingsystems}, opinions \cite{santurkar2023opinions}, values \cite{sorensen2024value}, and recommendations \cite{lyu-etal-2024-llm} fit within the content personalization umbrella. \textbf{Presentation Personalization} refers to "how" the LLM responds.  Style \cite{neelakanteswara-etal-2024-rags}, personality \cite{jiang-etal-2024-personallm}, formatting \cite{li-etal-2024-dissecting}, and verbosity \cite{hu2024explaininglengthbiasllmbased} all fit into presentation personalization.  A challenge when dealing with pairwise preferences is that, without feedback, it is difficult to know whether the content or presentation informed the user's choice.

\paragraph{Personalized Reward Models} Though many forms of steerable pluralism exist, we focus our main discussion on other personal reward model approaches.  Other forms of steerable pluralism are discussed in Appendix \ref{sec:deep-dives}.
We highlight differences in popular personal reward models in Table~\ref{tab:method-comparison}. Personalized reward models have been explored in the context of recommendation systems and learning from user interactions \cite{zhang2024provably, maghakian2023personalized}.  In NLP, other works have designed personal reward models for steerable alignment of LLMs. \citet{sorensen2024roadmap} discuss multi-objective reward modeling, where a reward model is made to balance several distinct objective functions.  Recent work has implemented this through training several reward models, and averaging either the outputs \cite{ramé2023rewarded} or the weights \cite{jang2023personalized, ramé2024warm}.  \citet{10.5555/3692070.3694392} finds that training language models with these multiple rewards in context enables steering at inference time.  Such methods require designated reward objectives defined a priori and thus have constrained preference functions.  In contrast, \texttt{SynthesizeMe} requires no such scaffolding.

\paragraph{Personalized Modeling from Interactions} A few methods exist outside of the scope of the axes-grounded reward models. \textbf{Group Preference Optimization} (GPO)~\cite{zhao2023grouppreferenceoptimizationfewshot} is a method to train a transformer for predicting group preferences from embeddings of prior preferences.  \textbf{Variational Preference Learning} \cite{poddar2024personalizingreinforcementlearninghuman} (VPL) takes several labelled user interactions as context and learns a user-specific embedding upon which to condition a reward model. \textbf{Pluralistic Alignment Framework} (PAL) \cite{chen2024pal} uses user interactions to learn a user weight over a finite set of preference prototypes to craft a personalized reward model. Concurrent to our work, \textbf{Fewshot Preference Optimization} (FSPO) \cite{singh2025fspofewshotpreferenceoptimization}, introduces a meta learning algorithm to fit personalized models on fewshot preferences with a persona generation step, and show that training a model on synthetic data helps generalize to real users.  \citet{zhang-2024-guided} introduces \textbf{Guided Profile Generation} (GPG), the work most conceptually similar to ours. GPG uses LLMs to generate specific user profiles based on their history to predict future preferences in product purchases and social media posts. Although conceptually similar, GPG operates within a constrained preference space, such as asking: ``\textit{Among the usage of 1. Capitalization, 2. Emoji, 3. Abbreviation, 4. Punctuation, which is the most distinctive feature of the above tweets?}".  In contrast, our work implicitly discovers personas without predefined constraints, allowing it to explore a broader preference space.

% Still to incorporate cleanly: \cite{fan-etal-2023-nano, maghakian2023personalized, zhang2024provably, salemi2024lamp, ramé2023rewarded, jang2023personalized, bakker2022finetuning, yuan2024selfrewarding, ramé2024warm, swamy2024minimaximalist, lambert2024alignment, casper2023open}

% \paragraph{Active Learning Human Preferences}
% Several prior works have explored adapting RL agents based on human-preferred trajectories \cite{biyik2018batch, biyik2019asking, ren2022efficient}.  Two main prior methods of query selection in active learning involve reducing the volume of the remaining search space \cite{sadigh2017active, palan2019learning, ren2022efficient}.  \cite{biyik2019asking} propose using information gain for query selection to learn from human preferences.  We adapt the RL active learning algorithm from \cite{biyik2019asking} to LLMs in this paper.  \citet{muldrew2024active} implement active learning for LLMs to select optimal pairs for direct preference optimization. However, they don't apply this method to individual personalization of LLMs.  To our knowledge, this is the first work on individual preference optimization for LLMs using efficient pairwise proposal selection.

\section{Introducing \texttt{SynthesizeMe}}
\label{sec:synthesize-me}
\subsection{Problem Formulation}
\label{subsec:problem-formulation}

Let $\mathcal{U}$ denote a population of users. Each user $u \in \mathcal{U}$ is associated with an unknown latent reward function
$R_u : \mathcal{T}_q \to \mathbb{R}$ where $\mathcal{T}_q$ represents the space of candidate responses $\tau_q^i$ to query $q \in \mathcal{Q}_u$. The function $R_u(q, \tau_q^i)$ quantifies the intrinsic utility that user $u$ assigns to response $\tau$.

\paragraph{Observations}

Rather than observing $R_u$ directly, we collect pairwise preference data:
\[
\mathcal{D}_u = \bigl\{ (\tau_q^{(1)}, \tau_q^{(2)}, y_q) \mid q \in \mathcal{Q}_u\}
\]
where
\[
y_q = \operatorname{sign}\Bigl( R_u\bigl(q,\tau_q^{(1)}\bigr) - R_u\bigl(q,\tau_q^{(2)}\bigr) \Bigr).
\]
Here, $y_q = +1$ indicates that $\tau_q^{(1)}$ is preferred over $\tau_q^{(2)}$, and $y_q = -1$ indicates the reverse.

\paragraph{Personalized Reward Modeling}

Our objective is to learn a personalized reward model $\hat{R}_u : \mathcal{T}_q \to \mathbb{R}$ that approximates $R_u$. We introduce a global configuration $\Omega$, which may include model parameters, prompt templates, or other alignment strategies to achieve this. $\Omega$ can be adapted to each user based on a small context set of pairwise comparisons, denoted $\mathcal{D}_u^{\text{context}}$, as follows: 
\[
\hat{R}_u = \operatorname{Adapt}\bigl( \mathcal{D}_u^{\text{context}} ; \Omega \bigr).
\]

\paragraph{Evaluation}
Performance of $\hat{R}_u$ is evaluated on a target set $\mathcal{D}_u^{\text{tgt}}$ by computing the pairwise accuracy.  We measure the fraction of comparisons where the model correctly predicts the user’s preference.

% A key insight of this work is as we collect $\left<\mathcal{P}(\tau_{wi}) > \mathcal{P}(\tau_{li})\right>$ in an online fashion we can use this to inform our future predictions.

\subsection{Method}
Key challenges for this setting include (1) data scarcity, as $|\mathcal{D}_u^{\text{context}}|$ is often between 5 to 15 pairs;  (2) preference attribution centers on understanding why a user picked a particular preference, and (3) overfitting to a limited set of preferences. To tackle these issues, we introduce \texttt{SynthesizeMe}, 
which tackles the low data challenge by extrapolating personas from limited interaction data and further proposing hypotheses about users' underlying preferences from their interactions.  To tackle the preference attribution challenge and validate these hypotheses \texttt{SynthesizeMe} uses a subset of $D_u^{\text{context}}$ as a validation set and only retains hypotheses which lead to improved validation accuracy.
Figure \ref{fig:icrm-method} showcases the \texttt{SynthesizeMe} pipeline.

\paragraph{Bootstrap Subroutine} We present \textsc{Bootstrap} in Algorithm \ref{alg:bootstrap_demos}.  In this procedure, we provide the LLM with the user's prompt and two model completions, asking it through Chain-of-Thought to: (1) explain which completion the user might prefer and why and (2) ultimately predict the user's preference. After the LLM makes a selection, we discard cases where it selects incorrectly. Optionally, we add context about the user to help the LLM reason.% This rejection sampling of reasoning is similar to Quiet-STaR \cite{zelikman2024quietstar} and specifically BootstrapFewshot from DSPy \cite{khattab2024dspy}.

\begin{algorithm}[t]
\caption{Bootstrap Reasoning + Demos}\label{alg:bootstrap_demos}
\begin{algorithmic}[1]
\Procedure{Bootstrap}{$\mathcal{D}_u^{\text{train}},\, \text{ctx}$}
  \State $\mathcal{R} \gets \varnothing$ \Comment{Set of successful reasoning}
  \ForAll{$(q,\tau_1,\tau_2,y) \in \mathcal{D}_u^{\text{train}}$}
    \State $l \gets$ \Call{LLMPredict}{$q,\, \tau_1,\, \tau_2,\, \text{ctx}$}
    \If{$\operatorname{sign}(l.\text{pred}) = y$}
      \State $\mathcal{R} \gets \mathcal{R} \cup \{\, (l.\text{rsn},\, (q,\tau_1,\tau_2,y)) \,\}$
    \EndIf
  \EndFor
  \State \Return $\mathcal{R}$
\EndProcedure
\end{algorithmic}
\end{algorithm}

\paragraph{Step 1. Bootstrap Reasoning} First, we prompt the model to generate reasoning for a set of pairwise preferences. %  Our bootstrap procedure is based on BootstrapFewshot from DSPy \cite{khattab2024dspy}.  We generate reasoning for pairwise preferences and reject any reasoning traces that lead to incorrect solutions.  
Notably, we assume no background knowledge of the user (context = $\varnothing$) at this stage of the pipeline, so the reasoning produced is purely speculation.  We use random subsets of the user's training preferences $\mathcal{D}_u^{\text{train}}$, evaluating on their validation preferences $\mathcal{D}_u^{\text{val}}$ for a fixed number of trials $n$. In practice, we use $n=10$. We then reject reasoning traces that improve prediction on the validation set.  This step can be described with the following expression.
\[
\argmax_{i \in \{1,\dots,n\}} \; \textsc{Eval}\Bigl( \textsc{Bootstrap}(\mathcal{D}_u^{\text{train}},\varnothing)_i,\; \mathcal{D}_u^{\text{val}} \Bigr).
\]

\paragraph{Step 2. Synthesize Persona} Using the validated reasoning about users, we synthesize a persona for each user. 
We take the bootstrapped reasoning and prior preferences from step 1 as contextual input \(\mathcal{R}^*\). Then, given a prompt \(\Theta\), we synthesize a user persona $\pi$ through a single call to an LLM:
\[
    \pi = \textsc{SynthesizePersona}(\mathcal{R}^*,\Theta).
\]
We optimize prompt $\Theta$ with a procedure described below. We find that optimized $\Theta$'s transfer well between models and preference datasets (\S\ref{subsec:transfer}), meaning that \texttt{SynthesizeMe} works for new user data and models without further optimization.  We show both the original and optimized prompts in Appendix~\ref{sec:prompt-optimization}.

\paragraph{Step 3. Extract Informative Examples} Finally, we leverage \(\pi\) as context to bootstrap and select the most informative demonstrations with $m$ trials:
\[
\argmax_{j \in \{1,\dots,m\}} \; \textsc{Eval}\Bigl( \textsc{Bootstrap}(\mathcal{D}_u^{\text{train}},\pi)_j,\; \mathcal{D}_u^{\text{val}} \Bigr).
\]
The persona \(\pi\) and demonstration set \(\mathcal{R}'^*\) are then used to personalize the reward model.  In practice, we use 10 trials ($m=10$).

\paragraph{Optimizing $\Theta$} We optimize prompt $\Theta$ using the DSPy MIPROv2 optimizer \cite{opsahl-ong-etal-2024-optimizing}, which rewrites user instructions and finds optimal demonstrations.  The outcome of our optimization is a natural language description of how to write a persona, alongside demonstrations of useful personas written for users in our trainset $\mathcal{U}_{train}$.  We only run our optimization on PRISM and find the optimized $\Theta$ transfers well to Chatbot Arena.  We include details of our prompt optimization and examples of optimized prompts in Appendix \ref{sec:prompt-optimization}. 

\section{Constructing PersonalRewardBench}
\label{sec:dataset}

To measure the adaptability of our personalized reward models to realistic settings, we use two existing datasets that provide per-user preference labels. \textbf{Chatbot Arena} \cite{zheng2023judging} is a dataset of $33,000$ in-the-wild conversations from $13,383$ users.  Users are tasked with judging the output of two distinct LLMs without knowing the model's identity through user-initiated conversations. \textbf{PRISM} \cite{kirk2024PRISMdataset} is a globally diverse preference dataset with a special focus on values and controversial opinions.  Users initiated 5 or 6 conversations with various LLMs on the platform. Unlike Chatbot Arena's pairwise preferences, these are $N$-way multi-turn preference sets.  We collect all $N \choose 2$ comparisons per turn to form our pairwise dataset.  Users also rate completions from 1 to 100, and we remove pairs where the users indicate less than 10\% difference in quality.

% We filter Chatbot Arena to users with five or more preference selections, stratify users by their number of preference pairs, then split into train, dev and test users with a random $40/10/50$ split. We also divide each user's preferences into context vs target preferences and further divide the target preferences into a dev and test set.  For this, we use a $50/20/30$ split.  Context preferences are used to learn about the user and predict their target preferences.

%After filtering to users with five or more preference selections, we are left with $1,004$ users.  The average number of preference pairs of the remaining users is $10.05$, and the median is $6$.  We split the users into train, dev, and test users to support our personalized preference exploration.  We stratify users by their number of preference pairs, so each split will have roughly the same distribution. Then, we perform a random $40/10/50$ split of users into train/dev/test sets.  Ultimately, this yielded $344$ train users, $98$ dev users, and $562$ test users.  We also divide each user's preferences into context vs target preferences and further divide the target preferences into a dev and test set.  For this, we use a $50/20/30$ split.  Context preferences are used to learn about the user and predict their target preferences.

\begin{table}[]
\centering
\resizebox{0.8\linewidth}{!}{%
\begin{NiceTabular}{@{}lcc@{}}[colortbl-like]
\toprule
\Block[fill=gray!20]{1-1}{} & 
\Block[fill=gray!20]{1-1}{\textbf{Chatbot Arena}} & 
\Block[fill=gray!20]{1-1}{\textbf{PRISM}} \\ \midrule
Users & 131 & 723 \\ \midrule
Median Conversations & 7 & 5 \\ 
Total Conversations & 1,338 & 3,897 \\ \midrule
Median Preference Pairs & 7 & 22 \\
Total Preference Pairs & 1,338 & 16,705 \\ \midrule
Median Unique Queries & 6 & 14 \\
Total Unique Queries & 1,170 & 10,935 \\ \midrule
MultiTurn (\%) & 14.65\% & 91.56\% \\
\bottomrule
\end{NiceTabular}%
}
\caption{Statistics of our filtered datasets comprising PersonalRewardBench.  After filtering for personalization, PRISM is much larger than Chatbot Arena. \label{tab:datasets}}
\end{table}

Not all Chatbot Arena and PRISM data are compatible with personalization. We devise a data filtering pipeline to get the highest-quality, most challenging, and most personalizable user data for benchmarking personal reward models.  Our pipeline consists of three stages: a \textbf{User Filter} which limits to only users with 5 or more preference pairs; a \textbf{Personalizable Filter} which uses GPT4o-mini to rate user queries for personalization potential; and finally a \textbf{Quality/Consensus Filter} which limits to only preference pairs that 5 LLM-as-a-judge reward models have high disagreement on (suggesting that the examples are controversial or opinionated).  We include a detailed breakdown of our filtering process in Appendix \ref{sec:filtering}.  Final statistics about the benchmark are provided in Table~\ref{tab:datasets}.

%PRISM \cite{kirk2024PRISMdataset} is a globally diverse dataset of about $8,000$ conversations with $68,371$ preference ratings.  The dataset has a special focus on values and controversial opinions.  After filtering to users with five or more interactions, we have $1,294$ users remaining.  However, unlike chatbot arena, these are $N$-way multiturn preference sets.  We collect all $N$ comparisons per turn with the preferred completion resulting in a median of $40$ preference pairs per user coming from a median of $6$ conversations. Splitting the users similarly to chatbot arena we obtain $515$ train users, $129$ dev users, and $650$ test users. 

\begin{table*}[h]
    \centering
    \resizebox{\textwidth}{!}{%
    \begin{tabular}{lccc|ccc}
        \toprule
        \rowcolor[gray]{0.9}
         & \multicolumn{3}{c|}{\textbf{Chatbot Arena}} & \multicolumn{3}{c}{\textbf{PRISM}} \\
        \rowcolor[gray]{0.9}
        Model & Llama 3.2 3B & Llama 3.1 8B & Llama 3.3 70B & Llama 3.2 3B & Llama 3.1 8B & Llama 3.3 70B \\
        \midrule
        % =========================
        %   Pairwise Methods
        % =========================
        Random & 50.00\% & 50.00\% & 50.00\% & 50.00\% & 50.00\% & 50.00\% \\
        \midrule
        \rowcolor[gray]{0.85} \multicolumn{7}{l}
        {\textbf{In-Context LLM as a Judge}} \\
        \midrule
        \rowcolor[gray]{0.9} \multicolumn{7}{l}{\textbf{\footnotesize Baselines -- LLM as a Judge}} \\
        Default & 54.23 {\scriptsize $\pm$ 4.14\%} & 53.70 {\scriptsize $\pm$ 4.05\%} & 56.69 {\scriptsize $\pm$ 4.05\%} & 51.65 {\scriptsize $\pm$ 1.25\%} & 52.80 {\scriptsize $\pm$ 1.24\%} & 54.35 {\scriptsize $\pm$ 1.24\%} \\
        Demographics & --- & --- & --- & 54.95 {\scriptsize $\pm$ 1.24\%} & 54.06 {\scriptsize $\pm$ 1.24\%} & 53.89 {\scriptsize $\pm$ 1.24\%} \\
        Memory & 52.29 {\scriptsize $\pm$ 4.23\%} & 58.10 {\scriptsize $\pm$ 4.05\%} & 57.57 {\scriptsize $\pm$ 4.05\%} & 50.86 {\scriptsize $\pm$ 1.26\%} & 54.17 {\scriptsize $\pm$ 1.24\%} & 54.20 {\scriptsize $\pm$ 1.24\%} \\
        \midrule
        \rowcolor[gray]{0.9} \multicolumn{7}{l}{\textbf{\footnotesize \texttt{SynthesizeMe} – LLM as a Judge (Ours)}} \\
        Just Demos & 53.17 {\scriptsize $\pm$ 4.05\%} & 55.11 {\scriptsize $\pm$ 4.05\%} & \textbf{61.97} {\scriptsize $\pm$ 3.96\%} & 51.70 {\scriptsize $\pm$ 1.25\%} & 54.93 {\scriptsize $\pm$ 1.24\%} & \textbf{57.76 {\scriptsize $\pm$ 1.25\%}} \\
        Just Personas & 50.88 {\scriptsize $\pm$ 4.14\%} & 57.39 {\scriptsize $\pm$ 4.05\%} & 53.70 {\scriptsize $\pm$ 4.05\%} & 51.12 {\scriptsize $\pm$ 1.27\%} & 53.66 {\scriptsize $\pm$ 1.24\%} & 53.84 {\scriptsize $\pm$ 1.25\%} \\
        Personas + Demos & 52.46 {\scriptsize $\pm$ 4.14\%} & 57.92 {\scriptsize $\pm$ 4.05\%} & \textbf{61.97} {\scriptsize $\pm$ 3.96\%} & 51.52 {\scriptsize $\pm$ 1.26\%} & 53.30 {\scriptsize $\pm$ 1.24\%} & 56.99 {\scriptsize $\pm$ 1.25\%} \\
        Personas + Distill $\Theta$ & \textbf{54.75} {\scriptsize $\pm$ 4.14\%} & 59.15 {\scriptsize $\pm$ 4.14\%} & --- & \textbf{52.21} {\scriptsize $\pm$ 1.25\%} & 54.95 {\scriptsize $\pm$ 1.24\%} & --- \\
        Personas + Demos + Distill $\Theta$ & \textbf{54.75} {\scriptsize $\pm$ 4.14\%} & \textbf{61.62} {\scriptsize $\pm$ 3.96\%} & --- & 52.09 {\scriptsize $\pm$ 1.25\%} & \textbf{55.24} {\scriptsize $\pm$ 1.25\%} & --- \\
        \midrule
        % =========================
        %   Pointwise Methods
        % =========================
        \rowcolor[gray]{0.85} \multicolumn{7}{l}{\textbf{Finetuned Reward Models}} \\
        \midrule
        \rowcolor[gray]{0.9} \multicolumn{7}{l}
        {\textbf{\footnotesize Existing Personal RM}} \\
        GPO & 53.87 {\scriptsize $\pm$ 4.14\%} & 56.69 {\scriptsize $\pm$ 4.05\%} & 58.10 {\scriptsize $\pm$ 4.05\%} & 55.26 {\scriptsize $\pm$ 1.25\%} & 56.48 {\scriptsize $\pm$ 1.24\%} & 55.65 {\scriptsize $\pm$ 1.24\%} \\

        VPL & 56.69 {\scriptsize $\pm$ 5.81\%} & 54.93 {\scriptsize $\pm$ 5.71\%} & --- & 58.26 {\scriptsize $\pm$ 1.75\%} & 58.23 {\scriptsize $\pm$ 1.75\%} & --- \\
        
        PAL & 60.56 {\scriptsize $\pm$ 5.63\%} & 56.69 {\scriptsize $\pm$ 5.81\%} & ---& 56.81 {\scriptsize $\pm$ 1.75\%} & 54.23 {\scriptsize $\pm$ 1.73\%} & --- \\
        \midrule
        \rowcolor[gray]{0.9} \multicolumn{7}{l}{\textbf{\footnotesize Bradley-Terry Reward Model}} \\
        $\dagger$ Finetuned Reward Model & \textbf{69.01} {\scriptsize $\pm$ 5.28\%} & 68.31 {\scriptsize $\pm$ 5.46\%} & 71.48 {\scriptsize $\pm$ 5.11\%} & \textbf{61.66} {\scriptsize $\pm$ 1.70\%} & \textbf{64.29} {\scriptsize $\pm$ 1.73\%} & 63.50 {\scriptsize $\pm$ 1.73\%} \\
        \midrule
        \rowcolor[gray]{0.9} \multicolumn{7}{l}{\textbf{\footnotesize \texttt{SynthesizeMe} -- Reward Model (Ours)}} \\
        $\dagger$ FT RM + Personas & \textbf{69.01} {\scriptsize $\pm$ 5.46\%} & 67.25 {\scriptsize $\pm$ 5.46\%} & \textbf{72.18} {\scriptsize $\pm$ 5.28\%} & 61.53 {\scriptsize $\pm$ 1.75\%} & 63.11 {\scriptsize $\pm$ 1.70\%} & \textbf{64.03} {\scriptsize $\pm$ 1.70\%} \\
        $\dagger$ FT RM + Personas + Demos & 66.55 {\scriptsize $\pm$ 5.46\%} & \textbf{69.72} {\scriptsize $\pm$ 5.28\%} & \textbf{72.18} {\scriptsize $\pm$ 5.28\%} & 61.24 {\scriptsize $\pm$ 1.70\%} & 62.74 {\scriptsize $\pm$ 1.73\%} & 63.44 {\scriptsize $\pm$ 1.70\%} \\
        \bottomrule
    \end{tabular}
    }
    \caption{Comparison of LLM judges and Finetuned Reward Models on Chatbot Arena and PRISM.  Distill $\Theta$ refers to learning the persona generation prompt (See Appendix \ref{sec:prompt-optimization}). \texttt{SynthesizeMe} works best at scale. Our results show that personalization with \texttt{SynthesizeMe} improves preference prediction accuracy for LLM Judge significantly and Reward Models slightly -- leading to state-of-the-art results with the latter. Note, we do not evaluate VPL 70b and PAL 70b due to hardware constraints.  All results reported with 95\% bootstrapped confidence intervals.  $\dagger$ Finetuned Reward Models are trained on unfiltered Chatbot Arena and PRISM datasets as they do not need context preferences.}
    \label{tab:results}
\end{table*}

To split the users into train, validation, and test sets, we stratify on their number of preference pairs to ensure an even distribution of "high resource" and "low resource" users.  We split into 40\% train, 10\% validation, and 50\% test users. Specifically, Chatbot Arena and PRISM have 23/19/89 and 280/65/378 train/dev/test users, respectively.

\section{Experiments}
We test \texttt{SynthesizeMe} on our PersonalRewardBench dataset alongside several personalized reward model methods that also learn from pairwise interactions.  Across all methods, we test three models of varying scales: Llama-3.2-3B, Llama-3.1-8B, and Llama-3.3-70B \cite{grattafiori2024llama3herdmodels}. 

\subsection{Baselines}
We briefly describe all baselines that we benchmark against here (more details in Appendix \ref{sec:deep-dives}).

\paragraph{LLM as a Judge Baselines} For the \textbf{Default} setup, we simply show the LLM the prompt and two completions and ask it to reason with chain of thought \cite{10.5555/3600270.3602070} to pick a preference.  PRISM provides demographic details of its users, so to test against \textbf{Demographics}, we try the LLM as a Judge prompt from ``\emph{Can LLM be a Personalized Judge?}" \cite{dong2024llmpersonalizedjudge}.  Finally, for \textbf{Memory}, we try to faithfully emulate ChatGPT's memory by keeping a running list of user knowledge (memory) in order of prior interactions, which we extract via an LLM.  We prompt the LLM to write 1-5 insights about a user from each interaction and take all of these insights as context.  All prompts are provided in Appendix~\ref{sec:prompts}.

\paragraph{Bradley-Terry Reward Models} We produce a \textbf{finetuned reward model} for Llama 3B, 8B, and 70B by training low-rank adapters on all data not in the target set of the test users.  This reward model is not personalized but fits the data distribution of each dataset. % of PRISM and Chatbot Arena.

\paragraph{Existing Personal Reward Models}  We test against three existing personal reward model algorithms which learn from brief user context: \textbf{Group Preference Optimization (GPO)} \cite{zhao2023grouppreferenceoptimizationfewshot}, \textbf{Variational Preference Learning (VPL)} \cite{poddar2024personalizingreinforcementlearninghuman}, and \textbf{Pluralistic Alignment Framework (PAL)} \cite{chen2024pal}.  We include implementation details and method descriptions for the baselines in Appendix \ref{sec:deep-dives}.

\subsection{Methods} \paragraph{LLM as a Judge + \texttt{SynthesizeMe}} We test five ablations for \texttt{SynthesizeMe} induced prompts for an LLM as a Judge Reward Model. \textbf{Just Demos}: We use only step 3 to extract informative demonstrations. \textbf{Just Personas}: We generate personas using steps 1 and 2, but exclude step 3 (demonstrations). \textbf{Personas + Demos}: We run the whole pipeline with all 3 steps, which adds both personas and optimal demonstrations using a single model.  \textbf{Personas + Distill $\Theta$}: We run steps 1 and 2, but replace prompt $\Theta$ in step 2 with a prompt learned using a larger model, in this case Llama-3.3-70B-Instruct.  \textbf{Personas + Demos + Distill $\Theta$}: This is our full method as it should be used in the wild.  We release our optimized persona generation prompt $\Theta$ for future use in \texttt{SynthesizeMe} personalization.
 
\paragraph{Finetuned Reward Model + \texttt{SynthesizeMe}} For all the users for whom we produce \texttt{SynthesizeMe} prompts, we include these prompts in the training data when finetuning the scalar reward model. During evaluation, we also use \texttt{SynthesizeMe} prompts. We use the personas and demos generated by the same LLM that we fine-tune.  In practice, we generate \texttt{SynthesizeMe} prompts for 25,878 out of 43,532 train entries in the PRISM dataset and 4,169 out of 23,025 train entries in ChatbotArena, while the rest of the users have too little data for meaningful personalization.

\subsection{Experimental Setting}
\label{sec:experiments}

We perform sweeps on the validation users to select optimal architecture and hyperparameters to ensure a fair comparison between methods. We outline hyperparameter sweeps in Appendix \ref{sec:gridsearches}.  All experiments were run with Llama-3.2-3B, Llama-3.1-8B, and Llama-3.3-70B \cite{grattafiori2024llama3herdmodels} on 1-8 NVIDIA H100 GPUs.  Training of Llama-3.3-70B was done on 4 NVIDIA H200 GPUs.

\subsection{Results and Analysis}

Table~\ref{tab:results} presents the results of all baselines and methods on PersonalRewardBench.  % We discuss a few main insights here.

\paragraph{\texttt{SynthesizeMe} helps LLM as a Judge.}  We find that adding \texttt{SynthesizeMe} induced prompts improves LLM as a Judge performance by up to 4.4\% on Chatbot Arena and 3.41\% on PRISM.  Across all LLM-as-a-Judge settings \texttt{SynthesizeMe} induced prompts push baselines to top performing methods.

\paragraph{\texttt{SynthesizeMe} can supplement Finetuned Reward Models.} In 3 out of 6 configurations \texttt{SynthesizeMe} augmented fine-tuned reward models slightly outperform the default fine-tuned reward model, primarily on ChatbotArena.  However, these improvements fall within confidence intervals and, as such, we primarily recommend \texttt{SynthesizeMe} as a tool for in-context personalization of LLM-as-a-Judge.

\paragraph{Finetuned Reward Models are a strong baseline given enough data.}  We find that Bradley-Terry Reward Models trained to fit the specific data distribution of ChatbotArena and PRISM outperform all LLM-as-a-Judge approaches and existing personalization baselines.  This fine-tuning is possible because of the thousands of examples of general user interaction data in-distribution of these datasets.  If such data is available, it is a strong baseline to fine-tune a reward model for your distribution before augmenting with personalization features.

\paragraph{Demonstrations in context are crucial for personalization.} Across all six settings we find that the winning configuration of LLM Judge + \texttt{SynthesizeMe} includes \textbf{demos}.  Such demonstrations can provide subtle nuance towards user preferences that are not fully captured by the personas.

\paragraph{Interactions are more useful than demographics.}  We find that the methods which rely on user interaction history for personalization of future preferences (\texttt{SynthesizeMe}, GPO, VPL, PAL), fare better than LLM Judge + Demographics.  For instance, with Llama 70B on PRISM, we find a 3.87\% gap between \texttt{SynthesizeMe} LLM as a Judge and Demographic LLM as a Judge.  The existing Personal RM approaches that make use of context examples (GPO, VPL, PAL) also consistently outperform demographic baselines by as much as 4.17\%. If selecting between demographics and interaction history collection for personalization, the interaction history is more valuable.

\paragraph{Distilling reasoning to smaller models works well.}  In our distillation setting,  we learn the persona generation prompt $\Theta$ using a 70b parameter model and apply it to smaller models.  For our 3B and 8B models, this is the most performant form of \texttt{SynthesizeMe}.  For instance, with Llama8B on chatbot arena this introduces a 3.7\% improvement.  We test an even more extensive version of this prompt sharing in (\S \ref{subsec:transfer}).

\section{Robustness of \texttt{SynthesizeMe}} 
We showcase the robustness of our \texttt{SynthesizeMe} method through (1) Scale (\S\ref{subsec:scaling}), (2) Interpretability (\S\ref{subsec:interpretability}), and (3) Model Transfer (\S\ref{subsec:transfer})

\subsection{Scaling}
\label{subsec:scaling}

\paragraph{How \texttt{SynthesizeMe} scales with model sizes.}

The results in Table~\ref{tab:results} demonstrate the scaling of method performance versus the size of the underlying model across 3B to 70B models.  We include supplemental visuals in Figures \ref{fig:arena_scaling} and \ref{fig:prism_scaling}.  \texttt{SynthesizeMe} scales to a higher accuracy than all other methods on Chatbot Arena as model size increases.  Similarly, on PRISM, \texttt{SynthesizeMe} scales well to match the best personal reward model performance from 52\% to 55\% to 58\% accuracy as the model scales from 3B to 70B.  When used on reward models instead of as LLM as a Judge prompts, the scaling is dictated more by the performance of the underlying reward model.  This makes \texttt{SynthesizeMe} prompts more future-proof as a new next big LLM with better reasoning is likely to continue improving performance.

\paragraph{How \texttt{SynthesizeMe} scales with more data.}

\begin{figure}
    \centering
    \includegraphics[width=1.0\linewidth]{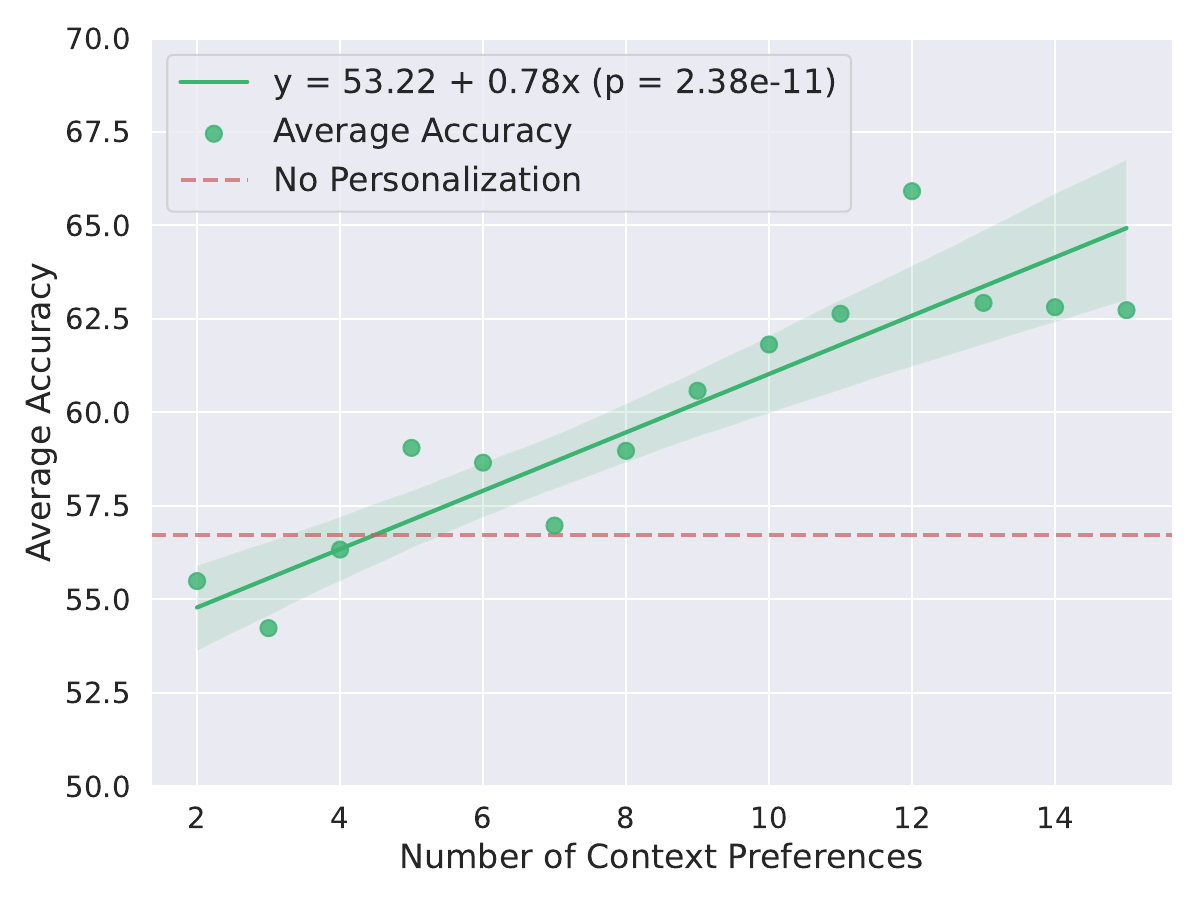}
    \caption{\texttt{SynthesizeMe} prompts for LLM-as-a-Judge scale well with increasing amounts of preferences per user on chatbot arena. We test with Llama-3.3-70B and find an almost 0.8\% improvement in accuracy for every additional context preference.  
    Just five context preferences beat non-personalized LLM as a Judge.}
    \label{fig:data_scaling}
\end{figure}

In Figure~\ref{fig:data_scaling}, we plot the accuracy of \texttt{SynthesizeMe} versus the number of preference pairs a user supplies.  We filter the dataset to only users that have $N$ or more context preferences and for any users with more we randomly sample just $N$.  We scale from $N=2 ... 15$.  As a user has more data, the accuracy of \texttt{SynthesizeMe} prompts for LLM-as-a-judge increases, suggesting this method scales well as users continue to engage with a platform.  We find Chatbot Area accuracy improves by about 0.8\% per additional context preference. We test this trend on Chatbot Arena because PRISM users were limited to 6 conversations on more scaffolded subjects.  In this way, a user with 15 preferences on Chatbot Arena may truly have discussed 15 different things on the platform. In contrast, a PRISM user will likely have provided several preferences on the same topic.  We show PRISM user scaling results in Figure~\ref{fig:both_datasets_user_scaling} in the Appendix.

\subsection{Interpretability}
\label{subsec:interpretability}
\begin{figure}
    \centering
    \includegraphics[width=1.0\linewidth]{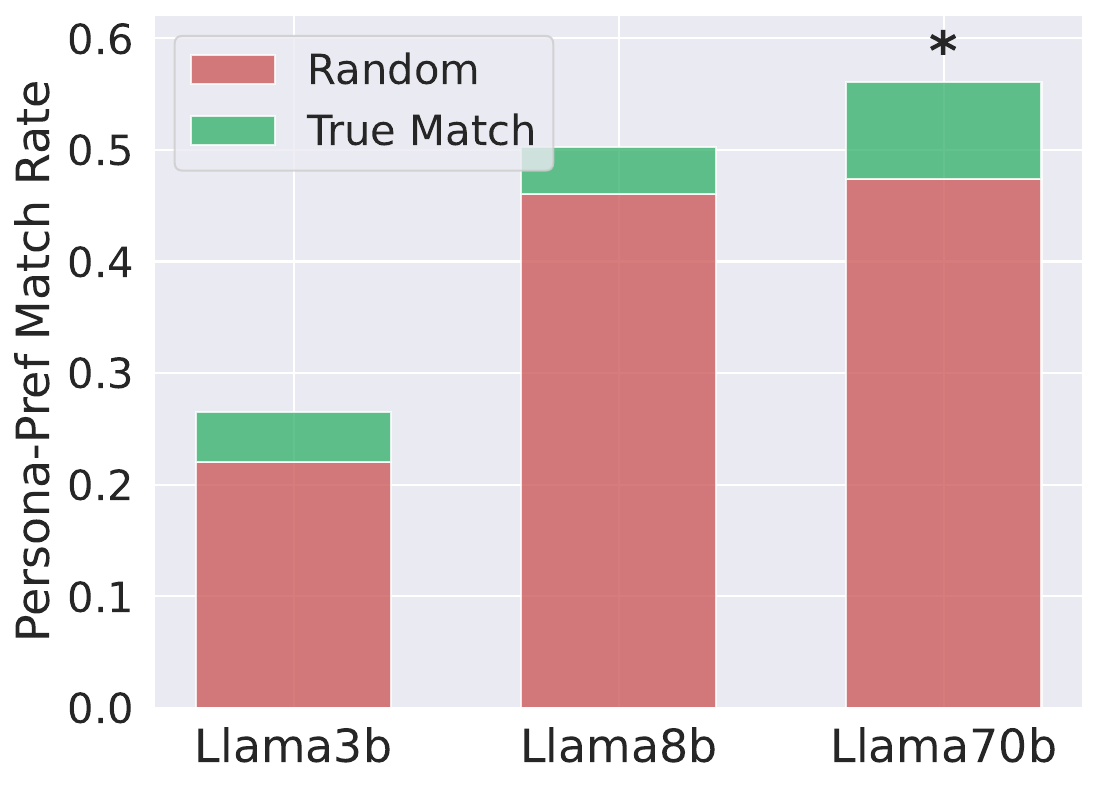}
    \caption{Rate at which personas generated by \texttt{SynthesizeMe} match the stated user preferences in PRISM.  We compare both the persona with the true user it came from and with a randomly selected user.  The true rate is always higher than the random rate, and for Llama 70b this holds with $p<0.05$.
    }
    \label{fig:match-rate}
\end{figure}

The final product of running \texttt{SynthesizeMe} is a personalized natural language prompt which can be ported from model to model to describe a user's personal preferences.  This prompt serves the same purpose as the user embeddings in VPL or PAL, but unlike large arrays of floating-point values, this prompt is far more interpretable.  Here we discuss two checks on the interpretability of the prompts.  We check (1) how faithful the prompts are to real users and (2) what these synthesized personas tell us about the users in PRISM and Chatbot Arena.

\paragraph{Validating the accuracy of synthesized personas} PRISM users provide 1-2 sentences on what they most want from an LLM at the start of onboarding to the platform.  Thus, we have ground truth labels for what users \textit{actually} care about.  In this experiment, we compare the synthesized personas from \texttt{SynthesizeMe} on PRISM with true user preferences.  We use GPT4o-mini to compare the user preference and persona to predict if they came from the same person (specifically, if they are a "strong match").  We report the DSPy Signature for this check in Figure~\ref{lst:signatures_match}.  To control for classification bias in this process, we measure the improvement of the match rate of the correctly paired personas with the match rate of random pairings.

Figure~\ref{fig:match-rate} shows the results of this check.  We find that as model size increases, the improvement over random grows more and more significant.  The rate of true matches increases from 26.5\% to 50.2\% from 3B to 8B.  The rate of matches increases from 50.2\% to 56.1\% from 8B to 70B while the rate of false positives remains relatively constant between 46-47\%.  In all cases, the learned personas match the user preferences in excess of random guessing.  In other words, these bootstrapped personas have reasonable overlap with users' actual preferences.

\paragraph{Learning about users from \texttt{SynthesizeMe} Personas}

Upon validating the personas, we turn to understanding more about the personas that comprise these datasets.  Using LLoom~\cite{lam2024conceptInduction}---a system for creating LLM-generated clusters of text-data---we synthesize $N=13$ clusters using the personas created by \texttt{SynthesizeMe}. We construct clusters using a corpus of personas from ChatbotArena. Generated clusters are fairly diverse (see Tab. \ref{tab:persona_counts}), ranging from users who prefer creative responses ($N = 35$) to users who prefer organized and analytical outputs ($N = 93$). Beyond broad preferences (e.g., creative vs. analytical), \texttt{SynthesizeMe} also produces personas that target \emph{particular} preferences of users. One cluster characterizes users who care about environmental concerns ($N=8$); another captures users who prefer humorous answers ($N=8$).  One could imagine using \texttt{SynthesizeMe} to not only better personalize for users but as a window into the true preferences and behavior of your users.

\begin{table}[t]
\resizebox{\linewidth}{!}{%
\centering
\small
\setlength{\tabcolsep}{5pt}
\begin{tabular}{l r l}
\toprule
\rowcolor[gray]{0.9}
\textbf{Cluster Name} & \multicolumn{2}{l}{\textbf{\;\;\;\# Personas}} \\[-0.7ex]
\midrule
Analytical Depth                  & \hphantom{0}93 & \PersonaBarCell{2.80} \\
Structured Information            & \hphantom{0}82 & \PersonaBarCell{2.47} \\
Curious Learner                   & \hphantom{0}79 & \PersonaBarCell{2.38} \\
Accuracy and Precision            & \hphantom{0}73 & \PersonaBarCell{2.20} \\
Balanced Perspectives             & \hphantom{0}60 & \PersonaBarCell{1.81} \\
Clear and Concise                 & \hphantom{0}46 & \PersonaBarCell{1.39} \\
Creativity Appreciation           & \hphantom{0}35 & \PersonaBarCell{1.05} \\
Emotional Communication           & \hphantom{0}26 & \PersonaBarCell{0.78} \\
Immersive Experiences             & \hphantom{0}26 & \PersonaBarCell{0.78} \\
Rich Storytelling                 & \hphantom{0}23 & \PersonaBarCell{0.69} \\
Practical Advice                  & \hphantom{0}22 & \PersonaBarCell{0.66} \\
Humor and Playfulness             & \hphantom{00}8 & \PersonaBarCell{0.24} \\
Social \& Environmental Concerns  & \hphantom{00}8 & \PersonaBarCell{0.24} \\
\bottomrule
\end{tabular}}
\caption{\texttt{SynthesizeMe} generates a diverse set of personas. We clustered personas from ChatbotArena, using LLoom \cite{lam2024conceptInduction}. Personas range from individuals who care about structured and analytic outputs to those who prioritize balanced perspectives or creativity.}
\label{tab:persona_counts}
\end{table}

% We include more details of our persona clustering in Appendix \ref{} \mr{@Omar we can add any interesting insights here! Maybe we'll end up putting the plot in the appendix?  Can move up with the 9 page camera ready limit.}

\subsection{Transferability}
\label{subsec:transfer}
From Table~\ref{tab:results}, we found that distilling the synthetic persona generation prompt, $\Theta$, from a more capable LLM to another is highly effective.  Here we instead test when entire sets of \texttt{SynthesizeMe} prompts are learned on top of one teacher model, then used to predict unseen preferences by another student model.  With this setup, practitioners could pay an upfront cost to compute a \texttt{SynthesizeMe} prompt for users with a large model, but from then on operate the actual reward model for cheaper.  Alternatively, one could save by learning the prompts with a cheaper model and transferring to a more expensive reward model.  We test both conditions.

\begin{figure}
    \centering
    \includegraphics[width=1.0\linewidth]{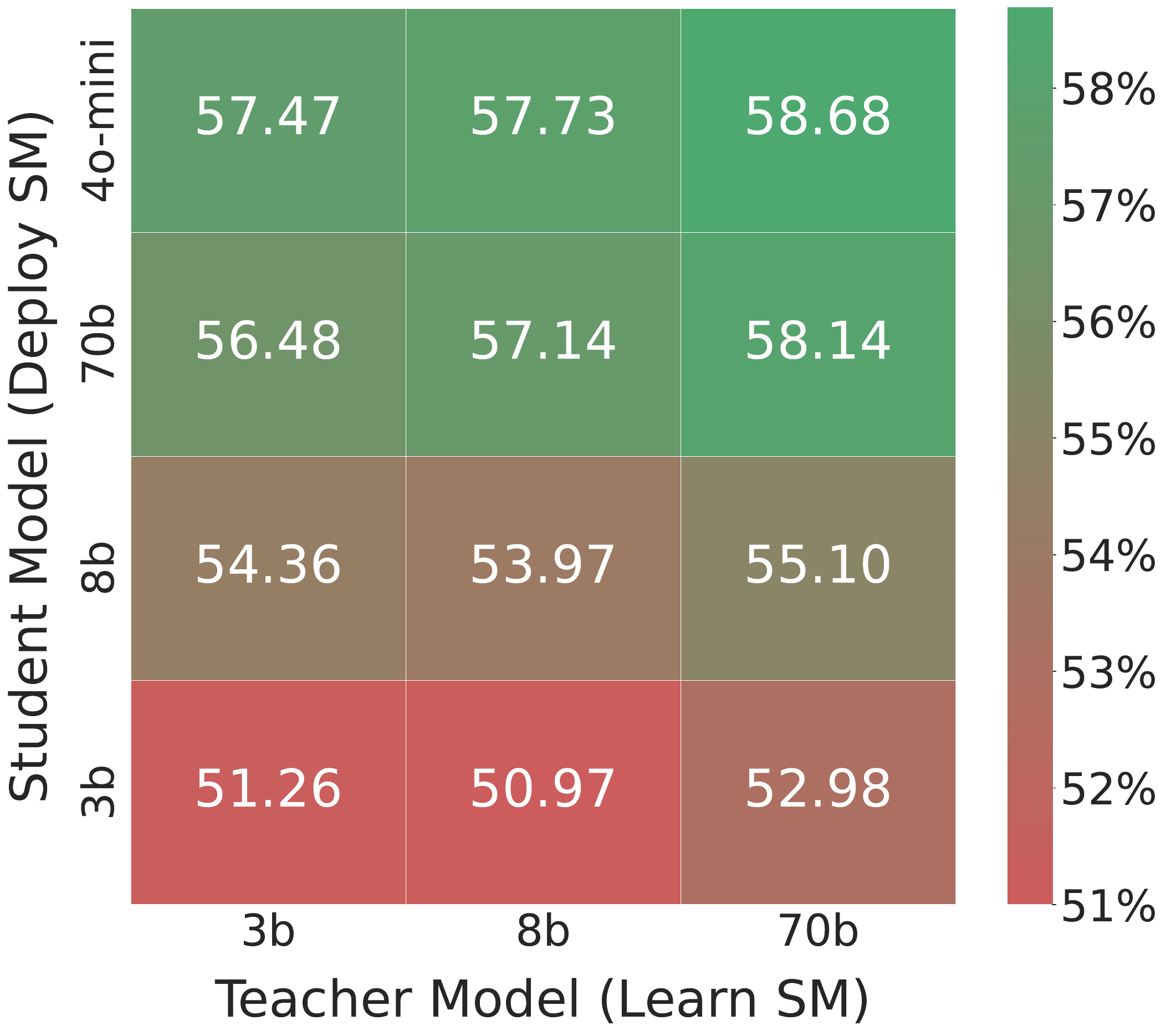}
    \caption{Results of transferring \texttt{SynthesizeMe} prompts learned on one model and testing on another.  GPT4o-mini works best and even personalizes on prompts learned by Llama 3.2 3B. % Larger than the 3b student model, prompts are fully compatible across models.  The largest effect comes from using the best student model. Learning prompts on more capable models leads to better distillation performance, however surprisingly a sufficiently capable large model can use prompts learned on a 3B model and still improve performance.
    }
    \label{fig:transfer-prism}
\end{figure}

Figure~\ref{fig:transfer-prism} presents a heatmap where the horizontal axis is the model used for training and the vertical axis is the model used for deployment.  We analyze results on PRISM, but Chatbot Arena was similar (See Figure~\ref{fig:transfer-arena}). As expected, larger student models are the biggest predictor of performance.  However, somewhat surprisingly, we find that weaker models can learn sufficient prompts to improve the personalization of larger models.  Overall, prompts transfer well between model sizes above 3B.

\paragraph{SynthesizeMe works across model families.} In Table~\ref{tab:other_models} we test SynthesizeMe (LLM as a Judge) versus default LLM as a Judge on Qwen, GPT, and Gemini family models.  For Qwen models, we report 95\% confidence intervals over 5 runs, while due to costs, we only report single-run results for GPT and Gemini models.  In 12 out of 14 conditions, \texttt{SynthesizeMe} improves over the default LLM as a Judge. In our multiple trial runs, it always improves over the baseline and does so with greater than 95\% confidence in 3 out of 6 conditions.  The only case where SynthesizeMe performs worse is on ChatbotArena with the latest Gemini models (Gemini-2.5).  Even in this case, SynthesizeMe is still more performant on PRISM.  As of publication, Gemini-2.5-Pro currently tops the Chatbot Arena leaderboard, and given that the Chatbot Arena data used in our study is publicly released, it is plausible that the latest Gemini models would be tuned on this data, which may explain why direct prompting does so well in this setting.

\begin{table}[t]
  \centering
  \small
  \setlength{\tabcolsep}{6pt}
  \renewcommand{\arraystretch}{1.12}
  \resizebox{\linewidth}{!}{%
  \begin{tabular}{lcc|cc}
    \toprule
    \rowcolor[gray]{0.9}
        & \multicolumn{2}{c|}{\textbf{Chatbot Arena}} 
        & \multicolumn{2}{c}{\textbf{PRISM}} \\
    \rowcolor[gray]{0.9}
    Model & Default & \texttt{SynthMe} & Default & \texttt{SynthMe} \\
    \midrule
      Qwen3-8B        & 61.41 {\scriptsize $\pm$ 0.98\%} & \textbf{61.83} {\scriptsize $\pm$ 2.04\%}\;\;\;\; & 55.14 {\scriptsize $\pm$ 0.36\%} & \textbf{55.95} {\scriptsize $\pm$ 0.41\%}*\;\; \\
      Qwen3-30B-A3B   & 60.74 {\scriptsize $\pm$ 1.11\%} & \textbf{63.91} {\scriptsize $\pm$ 1.85\%}** & 56.32 {\scriptsize $\pm$ 0.35\%} & \textbf{57.37} {\scriptsize $\pm$ 0.44\%}** \\
      Qwen3-32B       & 62.22 {\scriptsize $\pm$ 1.49\%} & \textbf{64.68} {\scriptsize $\pm$ 2.38\%}\;\;\; & 56.22 {\scriptsize $\pm$ 0.33\%} & \textbf{56.74} {\scriptsize $\pm$ 0.40\%}\;\;\; \\
      \midrule
      GPT4o-mini       & 59.86\% & \textbf{61.80}\% & 56.07\% & \textbf{58.90}\% \\
      \midrule
      Gemini-2.0-Flash    & 63.20\% & \textbf{64.61}\% & 56.97\% & \textbf{57.80}\% \\
      Gemini-2.5-Flash    & \textbf{67.25}\% & 66.73\% & 56.66\% & \textbf{58.36}\% \\
      Gemini-2.5-Pro    & \textbf{68.13}\% & 66.37\% & 56.51\% & \textbf{57.76}\% \\
    \bottomrule
  \end{tabular}}
  \caption{LLM as a Judge Accuracy with and without \texttt{SynthesizeMe}.  In 12 out of 14 conditions, \texttt{SynthesizeMe} outperforms the default LLM as a Judge, and in cases with five trials, it significantly outperforms LLM as a Judge 3 out of 6 times.  \texttt{SynthesizeMe} works well on many model families.  * = $0.01<p<0.05$, ** = $p<0.01$ determined by permutation test.}
  \label{tab:other_models}
\end{table}

\section{Conclusion}
We introduce \texttt{SynthesizeMe}, an approach to inducing synthetic user personas from user interactions for personalized reward modeling.  \texttt{SynthesizeMe} generates reasoning to interpret user preferences, derives synthetic user personas from that reasoning, and filters informative prior user interactions to create personalized prompts for the user. 
\texttt{SynthesizeMe} tackles both the data scarcity and preference attribution problem in reward model personalization by leveraging verifiable signals from the reward modeling task. %  We find that \textbf{LLMs generating and testing hypotheses is an extensible pattern for low resource personalization setting}. Many of our users had extremely limited data (5-15 context preferences).  Our process of bootstrapping reasoning and verifying on a validation set inspired by Quiet-STaR \cite{zelikman2024quietstar} and DSPy \cite{khattab2024dspy} helped find generalizable reasons why users may have selected a particular preference pair.
We further demonstrated that the prompts generated by our model are interpretable (\S\ref{subsec:interpretability}), transferable (\S\ref{subsec:transfer}), and scale well (\S\ref{subsec:scaling}). Overall, personalized reward models remain an important direction for pluralistic alignment, and \texttt{SynthesizeMe} is one step towards more interpretable pluralism.
Future work on LLM Personalization could invest in building a large-scale dataset of longitudinal and realistic preferences where users continue returning to the platform and providing more feedback, evolving with the system over time. Such a Wildchat style \cite{zhao2024wildchat} dataset could enable exciting research into user interactions far beyond personalization. %  and is a critical next step.
% The key next step in LLM Personalization is to \textbf{build a bigger dataset}. Right now Chatbot Arena and PRISM offer a strong start. Ultimately however, 15 or even 30 interactions with a user is extremely limited compared to the available data that companies have to personalize to their users' preferences.  Academia needs a dataset of longitudinal and realistic preferences where users continue returning to the platform and providing more feedback, evolving with the system over time.  

\section*{Limitations}

\texttt{SynthesizeMe} requires pairwise preference feedback to use as a verifiable signal when producing reasoning for user preferences.  The most lightweight personalization mechanisms will be able to infer user preferences without the need for pairwise data at all, similar to how people can learn how to interact with one another simply from interactions themselves.

All of our testing was done in the low-resource data regime for personalization (typically fewer than 25 preference pairs per user).  A more realistic personalization setting is one in which companies will collect user data over the course of hundreds of conversations and train personal models based on these longer-term interactions.  We did not test our methods on this because such large-scale, real, personalization data does not yet exist in academia.

\section*{Ethical Considerations}
Deploying models tailored to individual preferences introduces significant risks if not rigorously evaluated. For recommender systems, the most studied technological instantiation of personalization, research has found that systems can systematically influence user ratings and opinions even through simple interfaces~\citep{cosley2003seeing}, leading to concerns of amplification of extreme views~\citep{whittaker2021recommender}.

Personalizing models with much more expressive outputs, such as LLMs, has significant risks in this regard. Transparency is a key aspect in mitigating this influence as it allows users to interpret and even intervene on the influences a personalized algorithm has on them, leading to higher trust and satisfaction with transparent recommender systems~\citep{transparent}. Furthermore, to reduce extremes, personalized reward models should be deployed in the wild alongside systems that evaluate response adherence to strictly enforced ethical guidelines and principles.

An ongoing concern with human-LLM interaction is the problem of anthropomorphism \cite{schaaff2024impactsanthropomorphizinglargelanguage} and sycophancy \cite{sharma2024towards}.  Or in other words, people attributing human characteristics to AI, and AI saying what people \textbf{want} to hear over the truth.  Personalization has the potential to exacerbate these harms.  AI models that learn to fit user preferences specifically will be easier to attribute human-like qualities to.  Research has shown that friends influence each other's speaking patterns \cite{doi:10.1177/0146167291174008}, and personalization will be a way of reflecting your speech preferences onto AI.  Furthermore, through personalization, AI will learn to say what you, as the user, most want to hear, amplifying the already noted issue of sycophancy in non-personalized models.

\section*{Acknowledgment}
This research is supported in part by grants from ONR grant N000142412532, and NSF grant IIS-2247357.  We thank members of the Stanford SALT lab for their feedback and input. In particular we'd like to acknowledge Hao Zhu, Caleb Ziems, Rose Wang, Sherry Xie, Yijia Shao, Vyoma Raman, Ella Li, Ryan Louie, Shenguang Wu, Chenglei Si, and Yanzhe Zhang for their feedback and discussion of this work. 

% \section{Authorship Statement}

% All authors collaborated on writing this paper. \\

% \textbf{Michael}: Implemented and optimized ICRMs, built Open Feedback collection tool, user-stratified the Chatbot Arena and PRISM dataset, adapted GPO implementation to new data.  \\

% \textbf{Aditri}: Implemented PAL from scratch based on \cite{chen2024pal}, with different embeddings and context variants, PAL experiments and grid search, PAL data adaptations for PRISM. \\

% \textbf{Daniel}: Implemented VPL by extending the work of \cite{poddar2024personalizingreinforcementlearninghuman} for use with general preference data. Produced downstream data manipulations to extend user-stratified Chatbot Arena, PRISM, and Open Feedback data for use with VPL or other VAE models. Implemented new VPL options including support for newer llama models, auto-regressive context sampling strategies, pooling strategies, and more.  

\raggedbottom 

%Authorship statement: At the end of your paper (after the 'Acknowledgments' section in the template), please include a brief authorship statement, explaining how the individual authors contributed to the project. You are free to include whatever information you deem important to convey. For guidance, see the second page, right column, of this guidance for PNAS authors (p. 12). We are requiring this largely because we think it is a good policy in general. This statement is required even for singly-authored papers, because we want to know whether your project is a collaboration with people outside of the class. Only in extreme cases, and after discussion with the team, would we consider giving separate grades to team members based on this statement.

% \paragraph{Linear Projection} I need to take time to come up with a more thorough explanation for this method if we choose to include it although it doesn't work super well on real preferences so we may end up scrapping it.  The basic idea is you cut the last layer off the reward model and try to learn a new head that is fully consistent with all the prior preference pairs using Metropolis Hastings sampling.  This is inspired by active learning literature but works better in lower dimensional space.

% \clearpage
\bibliography{custom}

\appendix

\clearpage

\section{Extended Related Works and Deep Dives}
\label{sec:deep-dives}

\subsection{Steerable Pluralism}
\citet{sorensen2024roadmap} motivate the idea of "pluralistic alignment": aligning LLMs to fit diverse preferences and perspectives.  They outline three types of pluralistic alignment: overton pluralism \cite{lake2024distributional, hayati2024farextractdiverseperspectives}, steerable pluralism \cite{hwang-etal-2023-aligning, li-etal-2024-steerability, sharma2024facilitatingselfguidedmentalhealth}, and distributional pluralism \cite{zhao2023grouppreferenceoptimizationfewshot}.

Our work fits the steerable pluralism framework: "steering" models to reflect particular groups, individuals, or values.  Prior work in steerable alignment and personalization has used conversation history and memory \cite{salemi2024lamp, zhuang2024hydramodelfactorizationframework,zhang2024llmbasedmedicalassistantpersonalization} to recall user-specific details and opinions.  Other work introduces demographic information to both reward model judges \cite{dong2024llmpersonalizedjudge} or the LLM itself \cite{hu2024quantifying} to cater to particular groups.  \citet{feng-etal-2024-modular} proposes modular pluralism, where they train models on community corpora and select the right model to show to a user.  \citet{li2024personalizedlanguagemodelingpersonalized} do RLHF with contextual user embeddings. Finally, DITTO \cite{shaikh2025aligning} steers a model by fine-tuning on demonstrated user feedback.  Instead of memory, demographics, or demonstrations, \texttt{SynthesizeMe} learns from prior pairwise preferences to extract user personas for personalization.

\subsection{Group Preference Optimization (GPO)}  \cite{zhao2023grouppreferenceoptimizationfewshot} is an algorithm designed for learning a personalized or group preference in just a few context examples.  To achieve this, GPO breaks the reward modeling process into two stages: (1) Embedding preferences and (2) In-context learning.

\paragraph{Embedding Preferences} First, GPO embeds all user preferences using an LLM and averages the embeddings across all tokens for an input question.  In the original paper, \citet{zhao2023grouppreferenceoptimizationfewshot} test on survey questions, and produce an expected distribution of answers for the target group as outputs.  To map to our pairwise preference setting, we retain the input embedding strategy but instead, map to a single output [0,1] to indicate if the first (0) or second (1) completion was preferred.  This process produces a series of embedded inputs $x_1,...,x_n$ and corresponding output vectors $y_1,...,y_n$.

\paragraph{In-Context GPO Transformer} Given the embedded inputs and corresponding outputs, GPO trains a transformer model to predict preferences on unseen inputs $x_{n+1}, ..., x_m$.  Training groups are set apart to train the transformer model, and their full preferences are used as training data.  The GPO transformer uses a hidden dimension of 128, with four heads and six layers.  No positional embeddings are included to make the transformer invariant to the input order.  The transformer is trained using the cross-entropy loss versus the expected output vectors.  At test time, the transformer model is given all of the context preferences ($x_1,...x_n$, $y_1,...,y_n)$ as well as the target inputs $(x_{n+1},...x_m$ which are padded with zeros for $y$, and the output of the transformer for each input embedding is taken to be $y_{n+1},...y_m$.  In this way, the GPO transformer has learned the group's preference function in context from the given preferences.  It is worth noting that the GPO transformer is an embedding to reward transformer, not a text-to-text transformer.  The transformer itself is independent of the LLM.

\subsection{Variational Preference Learning (VPL)} Variational preference learning seeks to achieve pluralistic preference alignment by learning a latent space of variables that inform user-specific preferences. When predicting preferred outputs, VPL conditions its reward model on this latent space. We extend and update the implementation by \cite{poddar2024personalizingreinforcementlearninghuman} to support general preference learning.

To learn a reward model, VPL takes as input several labeled preference pairs from prior interactions with users $u \in U$. VPL's learning goal is then to predict several target preference pairs for users $u \in U$. More formally, the VPL dataset $\mathcal{D}$ consists of a set of prompts $x^i$, each with multiple responses  $r^i_j$, concatenated into a set of states $S^i_j = [x^i, r^i_j]$. For each user $u \in U$, we make the assumption that there is some user-specific reward function guiding their preferences, and we ask each user to label preferred responses across $S^i, i=1,...,N$. In labeling, we denote the users preferred response at turn $i$ as $S^i_A$ and their rejected response at turn $i$ as $S^i_B$. With this dataset $\mathcal{D}$, we can now describe the VPL architecture: 

\paragraph{VPL Encoder} Several previous preference pairs $[S^i_A, S^i_B]$ are sampled for each user. Each state $S_{A/B}^i$ is encoded by a frozen pre-trained large-language model, \texttt{LLM}, primarily to reduce input sizes. Towards reducing input sizes without loss of useful downstream task information, an LLM encoder was chosen (as opposed to smaller sentence encoder) as it has been shown that LLM encodings contain sufficient information for downstream tasks \cite{bhatia2023tart}. Furthermore, we pre-compute and cache LLM context encodings in the VPL encoder, so the additional computational complexity associated with using a larger model is minimal relative to overall training. After encoding all preference pair states with the LLM, we obtain embedding pairs $[e^i_A, e^i_B]$ and labels $y_i$ (indicating preferred state if not following the convention $r_\phi^u(S^i_A) > r_\phi^u(S^i_B)$). These pairs are then run though a \texttt{PairEncoder} consisting of a simple $2$-layer neural network. All context pairs are then passed into a \texttt{SequenceEncoder} which combines information from all contexts using a self-attention mechanism followed by simple linear projections to produce the latent dimension mean $\mu$ and variance $\Sigma$. 

\paragraph{VPL Decoder} To train the reward model, a latent vector $z$ is sampled from the posterior multivariate Gaussian distribution parameterized by the aforementioned $\mu$ and $\Sigma$ (note that during evaluation $\Sigma$ is set to $0$ for consistency). Given a new pair of states $S_A'$ and $S_B'$, we encode these with another \texttt{LLM} encoder, and concatenate the states with our latent $z$ before passing them through a simple $2$-layer neural net \texttt{Decoder} to yield reward values $r_A', r_B'$ for each state. During reward model training, we use an \textit{unfrozen} LLM encoder, with LoRA attention adapters applied for efficient backpropagation.

\subsection{Pluralistic Alignment Framework (PAL)} 

 We implemented PAL (Pluralistic Alignment Framework for Learning from Heterogeneous Preferences) \cite{chen2024pal}. Similarly to VPL, the PAL algorithm aims to learn a reward model for pluralistic preference alignment by representing each user’s preferences in a latent vector. PAL, however, achieves this across a set of archetypal persona axes: PAL produces user latents across a set of learnable preference prototypes that capture fundamental preference profiles, with user-specific weights that capture each user's unique preferences. By representing user preferences as a weighted combination of these prototypes, PAL is capable of generalizing across different users while also maintaining individual personalization.

 More formally, the model uses a set of $K$ learnable preference prototypes $\mathbf{P} = [\mathbf{p}_1, \ldots, \mathbf{p}_K]$ and user-specific weights $\mathbf{W} = [\mathbf{w}^{(1)}, \ldots, \mathbf{w}^{(N)}]$ (where N is the number of users) to capture the preferences of multiple users by approximating each user's preferences as a weighted combination of these prototypes. For a given user $i$, we capture their preferences as an "ideal point", $a^{(i)}$, which is computed as a weighted sum of prototypes, 
\begin{equation*}
    a^{(i)} = \mathbf{P} \cdot \mathbf{w}^{(i)}
\end{equation*}
where $\mathbf{w}^{(i)}$ is the user-specific weight vector and  $\mathbf{w}^{(i)} \in \Delta^K$ with $\sum_{k=1}^K w_k^{(i)} = 1$ and $w_k^{(i)} \geq 0$. This ensures each user’s preference is a convex combination of the prototypes.

Let \( f_{\theta} \) denote a shared mapping function that projects responses into the same latent space as the prototypes. For each preference pair $(x_{chosen}, x_{rejected})$ in the dataset, the embeddings of the chosen and rejected responses, \( f_{\text{chosen}} \) and \( f_{\text{rejected}} \), are produced by passing the response embeddings through the shared mapping function. The reward for each response (chosen or rejected) is then computed as the dot product between the mapped response embedding and the user’s ideal point \( a^{(i)} \). This similarity measure captures how closely each response aligns with the user’s preferences. Formally, for a response embedding \( f_{\text{choice}} \) and ideal point \( a^{(i)} \), the reward is given by:
\begin{equation*}
    \text{reward} = f_{\text{choice}} \cdot a^{(i)}
\end{equation*}
A higher reward indicates that the response is more aligned with the user’s preferences, as defined by their ideal point. This dot product approach directly measures alignment without requiring explicit distance calculations, simplifying the model and focusing on maximizing similarity with the user-specific preference profile.

To align the model’s predictions with the user-labeled preferences, we calculate the difference between the rewards for the chosen and rejected responses, using this difference to compute BCE loss, encouraging the model to assign a higher reward to the chosen response over the rejected response.

\section{Hyperparameters}
\label{sec:gridsearches}

\textbf{\texttt{SynthesizeMe}} For \texttt{SynthesizeMe} we use greedy decoding across three scales of Llama models.  We use up to the total number of train preference pairs a user provides as the upper limit of bootstrapped reasoning examples plus 4 demonstrations without reasoning to be included in our \texttt{SynthesizeMe} prompts.  We optimize the persona generation prompt on our set of training and validation users with the MIPROv2 optimizer \cite{opsahl-ong-etal-2024-optimizing}.  MIPROv2 was used to improve the prompt and optimize fewshot demonstrations by finding personas that are predictive of our validation user preferences.  We host the Llama-3.1-70B-Instruct model for inference on \texttt{4 x NVIDIA A6000} GPUs, the Llama-3.3-70B-Instruct model for inference on \texttt{2 x NVIDIA H100} GPUs, and Llama-3.2-3B-Instruct and Llama-3.1-8B-Instruct each on \texttt{1 x NVIDIA H100} GPU.  We find that increasing the number of trials when bootstrapping reasoning improves accuracy at a tradeoff with runtime, so we set our budget at 10 trials.

\paragraph{GPO Fine-tuning} 
\label{sec:gpo-exp-details}
To modify GPO to work for real user pairwise preferences, we needed to modify the prompt embedded by the LLM. We use Llama-3.1-8B-Instruct, Llama-3.2-3B-Instruct, and Llama-3.3-70B-Instruct to produce our embeddings over an input prompt that contains the user context and both possible assistant completions. Then, the output of GPO is either a zero (indicating the first completion is preferred) or a one (for the second completion). We maintain the GPO-transformer hidden dimension of \{64, 128, 256\}, with \{2,4,8\} attention heads and \{4,6,8\} layers.  We train the GPO-transformer using the Adam optimizer with a learning rate of $3e-5$ and cosine annealing for 200000 steps.  We used mean-pooling or last layer for the embeddings.  Training was performed on \texttt{1 NVIDIA A6000} GPU.

\paragraph{VPL Fine-tuning}
\label{sec:vpl-exp-details}
 In our extension of the VPL work, we keep the architecture largely the same as the original paper and add several new hyperparameters to test various embedding pooling strategies and autoregressive context sampling strategies. We train VPL reward models using \texttt{llama-3.2-3b-instruct}, \texttt{llama-3.1-8b-instruct} as the base LLM encoder for both cached VPL encoder embeddings and LoRA fine-tuned VPL decoder embeddings. During training, we learned the entire model using the base log-sigmoid loss used by the original VPL paper to maximize chosen rewards relative to rejected rewards, along with a KL divergence loss on the Gaussian to regularize the model against a standard multivariate normal (as is standard for variational autoencoders). We utilize a learning rate of \texttt{3e-4}, AdamW \cite{loshchilov2017adamw} and BF16 float precision as per the original paper. In our initial VPL gridsearch we experiment with \texttt{mean} vs. \texttt{last}-token pooling for the LLM embeddings extracted from the last hidden state. We use random autoregressive context sampling (meaning we sample randomly from earlier in the current data point's conversation or other user conversations) with sample sizes of $5$, $10$, and $15$. In the second gridsearch with \texttt{llama-3.1-8b-instruct} we build upon the results from the first gridsearch, using last-pooling and testing random contexts with sample sizes $5$, $10$, and $15$.  Training performed on \texttt{1 x NVIDIA H100} GPU.

 \paragraph{PAL Fine-tuning} 
 \label{sec:pal-exp-details}
 For PAL, we use many of the same hyperparameters from the recommended settings in the original paper \cite{chen2024pal}.  We run PAL B with frozen LLM parameters and set the dimension of the preference embeddings equal to the hidden dimension of the LLM encoder (3072 for Llama 3.2 3B, 4096 for Llama 3.1 8B, and 8192 for Llama 3.3 70B though ultimately we did not fine-tune Llama 3.3 70B due to memory constraints).  We used a 2-layer MLP with gelu activations for the projection architecture with Gaussian initialization and disabled learnable temperature.  We set the number of prototypical points ($K$) to 2.  We train with batch size 2 for the 3072 dimensional model and batch size 1 for the 4096 dimensional model.  We sweep over using mean pooling or last token for the LLM encoding and find that last token consistently outperforms mean pooling on the validation set so we report last token throughout the main paper.

 \paragraph{Bradley Terry Reward Model Finetuning}

We include the hyperparameters used across all reward model fine-tuning runs (Llama 3.2 3B, Llama 3.1 8B, and Llama 3.3 70B) in Table~\ref{tab:rm-hparams}.  For all models, we train a rank 64 LoRA adapter using standard contrastive reward modeling loss.  We use the default reward model trainer from HuggingFace TRL \cite{vonwerra2022trl}.

\begin{table}[t]
  \centering
  \small
  \setlength{\tabcolsep}{6pt}
  \renewcommand{\arraystretch}{1.1}
  % \resizebox{\linewidth}{!}{
  \begin{tabular}{ll}
    \toprule
    \rowcolor[gray]{0.9}
    \textbf{Hyperparameter} & \textbf{Value} \\
    \midrule
    Per-device batch size        & 1 \\
    Training epochs              & 1 \\
    Learning rate                & $1\times10^{-5}$ \\
    Max sequence length          & 8192 tokens \\
    LoRA rank $r$                & 64 \\
    LoRA $\alpha$                & 32 \\
    LoRA target modules          & \texttt{q\_proj}, \texttt{v\_proj} \\
    Max gradient norm            & 1.0 \\
    Precision                    & bfloat16 \\
    Optimizer                    & Adam \\
    Adam betas $(\beta_1,\beta_2)$ & $(0.9,\;0.999)$ \\
    Weight decay                 & 0.0 \\
    Warm-up steps                & 0 \\
    \bottomrule
  \end{tabular}%}
  \caption{Shared hyperparameters for all reward-model fine-tuning runs.  
  Hardware differs by model size: 1 × A100 (3 B), 2 × A100 (8 B), and 4 × H200 (70 B).}
  \label{tab:rm-hparams}
\end{table}

\section{PersonalRewardBench Filtering}
\label{sec:filtering}
Here we discuss in detail our pipeline for filtering Chatbot Arena and PRISM. Examples of removed conversations and the amount of data filtered are available in Table \ref{tab:filtering}.

\paragraph{User Filter} We begin by stratifying the datasets by user and filtering to users with only five or more preference pairs.  This filters a huge chunk of Chatbot Arena (over 10,000 users), but retains much of PRISM.  Users with fewer than five preference pairs cannot be used for personalization as there isn't enough data to form a context and eval set.

\paragraph{Personalizable Filter}  Many queries, especially in Chatbot Arena, are not suitable for personalization.  Many users come to the platform to test and judge models, not to ask standard questions.  In fact, the most popular things users ask about in Chatbot arena are (1) software errors and (2) questions about AI and software \cite{zheng2024lmsyschatm}.  As such we use an LLM to filter to only queries and responses deemed "personalizable".  

We devised criteria to define personalizable queries; however, at the heart of the criteria is the question, "Would reasonable users potentially disagree about how this question should be answered?".  We manually labeled 100 examples from Chatbot Arena and prompted GPT-4o-mini with our criteria.  Using a 20/30/50 train/dev/test split we found performant fewshot demos which increased our accuracy of the LLM filter from 77\% to 83\%.  We ran on all data and filtered our conversations that did not meet the criteria for personalization.  We include the full criteria prompt in Figure~\ref{fig:personalizable-test-prompt}.  This filtering step has useful implications because it can potentially be used in LLM chat interfaces to decide whether or not to surface pairwise completions to the user for feedback.  If the query is not personalizable, there is less reason to collect feedback.

\begin{figure*}
\begin{tcolorbox}[colback=blue!5!white, colframe=blue!75!black, title=Personalization Filter Prompt (Pt. 1/3)]
\# Annotation Guidelines for Personalizable Queries and Responses \\

\#\# Query-Level Personalization \\

**\textbf{Personalizable Query:}**  \\
A query is personalizable if it invites variation or subjective interpretation. For example:\\

- **\textbf{Creative or stylistic queries}** (e.g., "Rewrite this text in the style of Hemingway").  \\
- **\textbf{Subjective questions}** (e.g., "What are the best books for a software engineer?").  \\
- **\textbf{Open-ended or speculative queries}** (e.g., "Describe a day in an alternate reality.").  \\
- **\textbf{Requests for summaries, descriptions, or explanations}** where the phrasing, depth, or focus can vary significantly based on user preferences (e.g., "Explain gravity to a 5-year-old" or "Tell me about the Onin Rebellion").  \\
- **\textbf{Technical or instructional queries}** that allow variation in explanation style, tone, or complexity (e.g., "How can I quantize a model to 4 bits?" or "Explain LSAT Question 5").  \\
- **\textbf{Creative programming tasks}** that can vary in code style, documentation, modularity, or clarity (e.g., "Write a Python program for X").  \\

**\textbf{Non-Personalizable Query:}**  \\
A query is non-personalizable if it:\\

- **\textbf{Tests the model’s capability}** without reflecting user preferences (e.g., toy tasks or trick questions like "Add one line of Java code" or "What is the etymology of 'glibbermood'?").  \\
- **Requests formatting or reorganization** of already provided content (e.g., "Reformat this into a list").  \\
- **\textbf{Has a single correct answer}** or asks for factual information (e.g., "What is 2+2?" or "List the demonyms for the 12 most populated Arabic countries").  \\
- **\textbf{Explicitly requests formatting tasks}** based on already given instructions (e.g., "Reformat these instructions for making tea into bullet points").  \\
- **\textbf{Directly asks for translations or exact outputs}** (e.g., "Translate this into German").  \\

---\\

\#\# Response-Level Personalization\\

**\textbf{Personalizable Responses:}**  \\
Responses are personalizable if they:\\

- Differ meaningfully in **\textbf{content, tone, style, or depth}**.  \\
- Reflect **\textbf{creative or subjective interpretations}** (e.g., "Two different short stories with distinct tones").  \\
- Show distinct approaches to answering the same query, even for factual topics, where the level of detail, focus, or presentation varies significantly.  \\
- Offer different emphases or priorities in explaining open-ended queries (e.g., focusing on cultural impact vs. military strategy in a historical event).  \\

**\textbf{Non-Personalizable Responses}:**  \\
Responses are not personalizable if they:

\end{tcolorbox}
\end{figure*}
\begin{figure*}
\begin{tcolorbox}[colback=blue!5!white, colframe=blue!75!black, title=Personalization Filter Prompt (Pt. 2/3)]

- Are **\textbf{factually incorrect or nonsensical}**. Logical inconsistencies render responses unsuitable for personalization, as the focus shifts to factual accuracy.  \\
- Differ only trivially, such as slight variations in phrasing.  
- Are inappropriately inconsistent with the query (e.g., responses in a different language without user intent).  \\
- Refuse to answer when refusal is inappropriate or inconsistent.  
- Reflect differences that focus solely on completeness or factual correctness without subjective variation.  \\

The right question to ask is: "Could two completely reasonable people disagree on which of these responses is better?" If that answer is **\textbf{YES}**, then it is probably a personalizable set of responses!\\

---\\

\#\# Gray Areas\\

1. **\textbf{Reasonable Disagreement:}** If people might reasonably disagree on the appropriateness of answering (e.g., ethical dilemmas), both the query and responses may still be personalizable if they reflect subjective or varying interpretations.  \\
2. **\textbf{Inconsistent Behavior:}** Responses that differ due to inconsistent model behavior (e.g., refusal vs. compliance) are not personalizable unless reasonable people would disagree on the necessity of refusal.  \\
3. **\textbf{Toy or Trick Queries:}** Queries designed to "test" the model (e.g., adding a single line of code, impossible tasks) are not personalizable. However, responses to such queries may still exhibit meaningful personalization if they vary significantly in tone, depth, or creativity.  \\
4. **\textbf{Formatting or Reorganization Requests:}** Queries that explicitly ask for information to be reorganized (e.g., "Reformat this into a list") are typically non-personalizable unless the responses exhibit significant variation in structure or additional creative input beyond the request.  \\
5. **\textbf{Open-Ended Summaries or Explanations:}** Queries that request general information (e.g., "Tell me about X") are often personalizable due to the wide range of potential angles, tones, and depths available to answer them. Assess whether responses demonstrate meaningful variation in these aspects.  \\
6. **\textbf{Logical Consistency in Responses:}** Logical inconsistencies or factual errors in responses detract from their personalization potential. Even if a query invites personalization, incorrect or incoherent responses are categorized as non-personalizable.  \\

---\\

\# Examples Section\\

\#\# **\textbf{Personalizable Queries and Responses}**  \\
1. **\textbf{Query:}** "Explain gravity to a 5-year-old."  \\
   - **\textbf{Personalizable Query:}** Yes  \\
   - **\textbf{Personalizable Responses:}** Yes, as explanations can vary in tone, creativity, and complexity.  \\

2. **\textbf{Query:}** "Write a Hemingway-style description of a beach."  \\
   - **\textbf{Personalizable Query:}** Yes  \\
   - **\textbf{Personalizable Responses:}** Yes, as responses can differ in their adherence to Hemingway's style.

\end{tcolorbox}
\end{figure*}

\begin{figure*}
\begin{tcolorbox}[colback=blue!5!white, colframe=blue!75!black, title=Personalization Filter Prompt (Pt. 3/3)]

3. **\textbf{Query:}** "Summarize the Bible."  \\
   - **\textbf{Personalizable Query:}** Yes  \\
   - **\textbf{Personalizable Responses:}** Yes, as summaries can emphasize theological, historical, or narrative elements.  \\

\#\#\#\# **\textbf{Non-Personalizable Queries and Responses}**  \\
1. **\textbf{Query:}** "What is 2+2?"  \\
   - **\textbf{Personalizable Query:}** No, as it has a single correct answer.  \\
   - **\textbf{Personalizable Responses:}** No, as differences only reflect correctness.  \\

2. **\textbf{Query:}** "Reformat these instructions into bullet points."  \\
   - **\textbf{Personalizable Query:}** No, as the task is purely formatting.  \\
   - **\textbf{Personalizable Responses:}** No, unless the responses provide creative restructuring beyond the query.  \\

3. **\textbf{Query:}** "Translate this into German."  \\
   - **\textbf{Personalizable Query:}** No, as it seeks a straightforward translation.  \\
    - **\textbf{Personalizable Responses:}** No, as variations are trivial.
    
\#\# **\textbf{Gray Area Examples}**  \\
1. **\textbf{Query:}** "Should the assistant help build an AI with specific characteristics?"  \\
   - **\textbf{Personalizable Query:}** Yes, as reasonable people may disagree on fulfilling the request.  \\
   - **\textbf{Personalizable Responses:}** Yes, if responses reflect ethical considerations and subjective preferences. \\ 

2. **\textbf{Query:}** "Why do chatbots use the phrase 'as an AI language model'?"  \\
   - **\textbf{Personalizable Query:}** Yes, as it invites reasoning and subjective interpretations.  \\
   - **\textbf{Personalizable Responses:}** Yes, if responses vary in tone and depth.  \\

3. **\textbf{Query:}** "Summarize Monte Carlo methods in reinforcement learning."  \\
   - **\textbf{Personalizable Query:}** Yes, as summaries can vary in technical depth and focus.  \\
   - **\textbf{Personalizable Responses:}** No, if one response is incorrect or lacks coherence.  \\

These examples provide practical clarity on when queries and responses should be considered personalizable. Use them as a reference for future annotations!\\

... [[15 Examples Selected through DSPy Optimization]] ...
\end{tcolorbox}
\caption{\label{fig:personalizable-test-prompt}Personalization Prompt used to filter out conversations that are not personalizable from the dataset}
\end{figure*}

\paragraph{Quality Filter}  Another issue we find in Chatbot Arena and PRISM is that these benchmarks tested LLMs of varying capabilities and scales.  The generation quality of some LLMs in the benchmark are not comparable to others.  Occasionally, in the completions, one model sufficiently answers the question, and the other outputs nonsense or answers an unrelated question.  The correct choice is clear in these cases, and there is no reason to use this data for personalization.  On the other hand, the user may make a mistake in their preference selection and choose nonsensical answers accidentally or adversarially.  
We introduce a quality filter based on simulated annotator disagreement to address both cases.  We run five popular models: GPT-4o-mini\footnote{\href{https://openai.com/index/gpt-4o-mini-advancing-cost-efficient-intelligence/}{GPT-4o-mini blog}}, Llama-3.1-70B-Instruct \cite{grattafiori2024llama3herdmodels}, Llama-3.3-70B-Instruct \cite{grattafiori2024llama3herdmodels}, Gemini 1.5 Pro \cite{geminiteam2024geminifamilyhighlycapable}, and Qwen2.5-72B-Instruct \cite{qwen2025qwen25technicalreport} as LLM-as-a-judge in both orderings of the possible completions for a total of 10 judgments.  We remove cases where all five models agree (100\% accurate) for being too obvious or adversarial.

\paragraph{Splits}   We first divide into 40\% train users 10\% validation users, and 50\% test users.  Then per user, we ordered their conversations temporally (to avoid leakage) and stored the first 50\% as context\_train preferences ($\mathcal{D}_u^{\text{train}}$), the next 20\% as context\_validation preferences ($\mathcal{D}_u^{\text{val}}$), and the final 30\% as target preferences ($\mathcal{D}_u^{\text{tgt}}$).  Ultimately, our user splits are 23/19/89 for chatbot arena\footnote{We produce these splits after step 1 in our pipeline, so the percentage of users in each split of chatbot arena varies slightly from the original distribution after additional filtering.} and 280/65/378 for prism. 

\begin{table*}[ht]
\centering
\resizebox{\textwidth}{!}{%
\begin{NiceTabular}{ccccp{0.55\textwidth}}[colortbl-like]
\toprule
\Block[fill=gray!20]{2-1}{\small\textbf{Step}} & 
\Block[fill=gray!20]{2-1}{\small\textbf{Operation}} & 
\Block[fill=gray!20]{1-2}{\small\textbf{Preference Pairs (Users)}} & &
\Block[fill=gray!20]{2-1}{\small\textbf{Example User Queries Removed}} \\
 &  & \Block[fill=gray!20]{}{\small \textbf{Chatbot Arena}} & \Block[fill=gray!20]{}{\small \textbf{PRISM}} &  \\ \midrule
0 & (Original) & 33,000 (13,383) & 68,371 (1,396) & -- \\ \midrule
1 & User Filter & 10,092 (1,004) & 52,580 (1,294) & \textit{"what is the 145th most popular language"} \\
2 & Personalizable Filter & 3,927 (353) & 26,663 (734) & \textit{"Please sort these numbers: 6, 4, 2, 7, 5, 11, 1"} \\
3 & Quality Filter & 1,338 (131) & 16,705 (720) & \textit{"Name films like the video game Factorio"} \\ \bottomrule
\end{NiceTabular}
}%
\caption{Amount of data after each step of our data filtering pipeline and example queries from removed conversations. \label{tab:filtering}}
\vspace{-3mm}
\end{table*}

\section{Additional Figures}

\begin{figure}[t]
    \centering
    \includegraphics[width=1.0\linewidth]{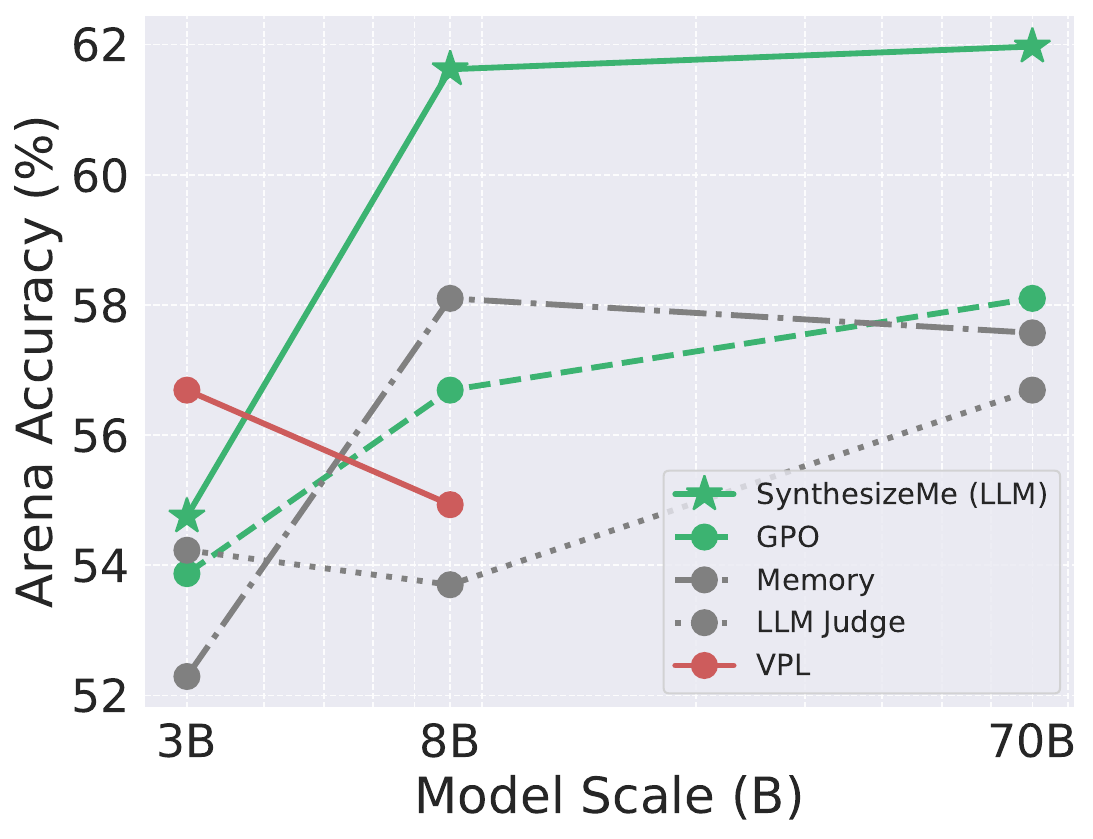}
    \caption{Scaling methods from Llama 3b to 70b on ChatbotArena.  Methods shown in green improve across scale, gray fluctuate, and red decrease with scale.  }
    \label{fig:arena_scaling}
\end{figure}

\begin{figure}[t]
    \centering
    \includegraphics[width=1.0\linewidth]{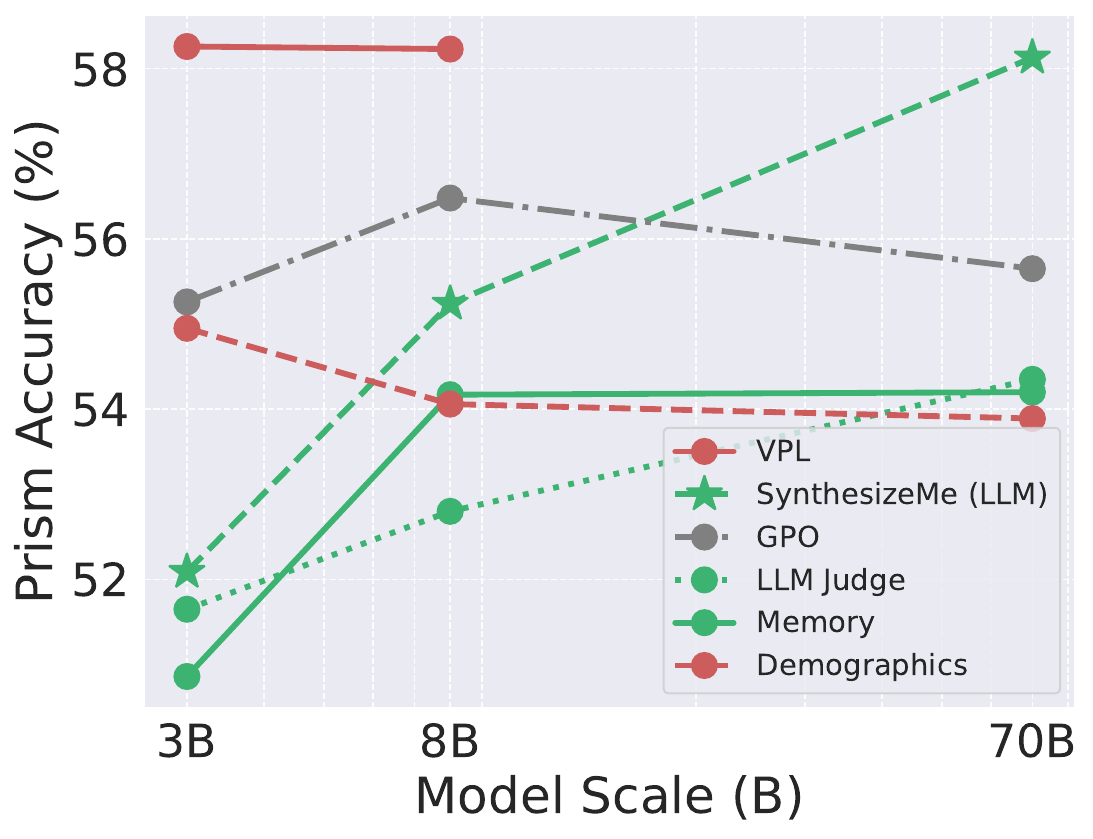}
    \caption{Scaling methods from Llama 3b to 70b on PRISM. Methods shown in green improve across scale, gray fluctuate, and red decrease with scale. }
    \label{fig:prism_scaling}
\end{figure}

\begin{figure}[t!] 
    \centering
    \includegraphics[width=1.0\linewidth]{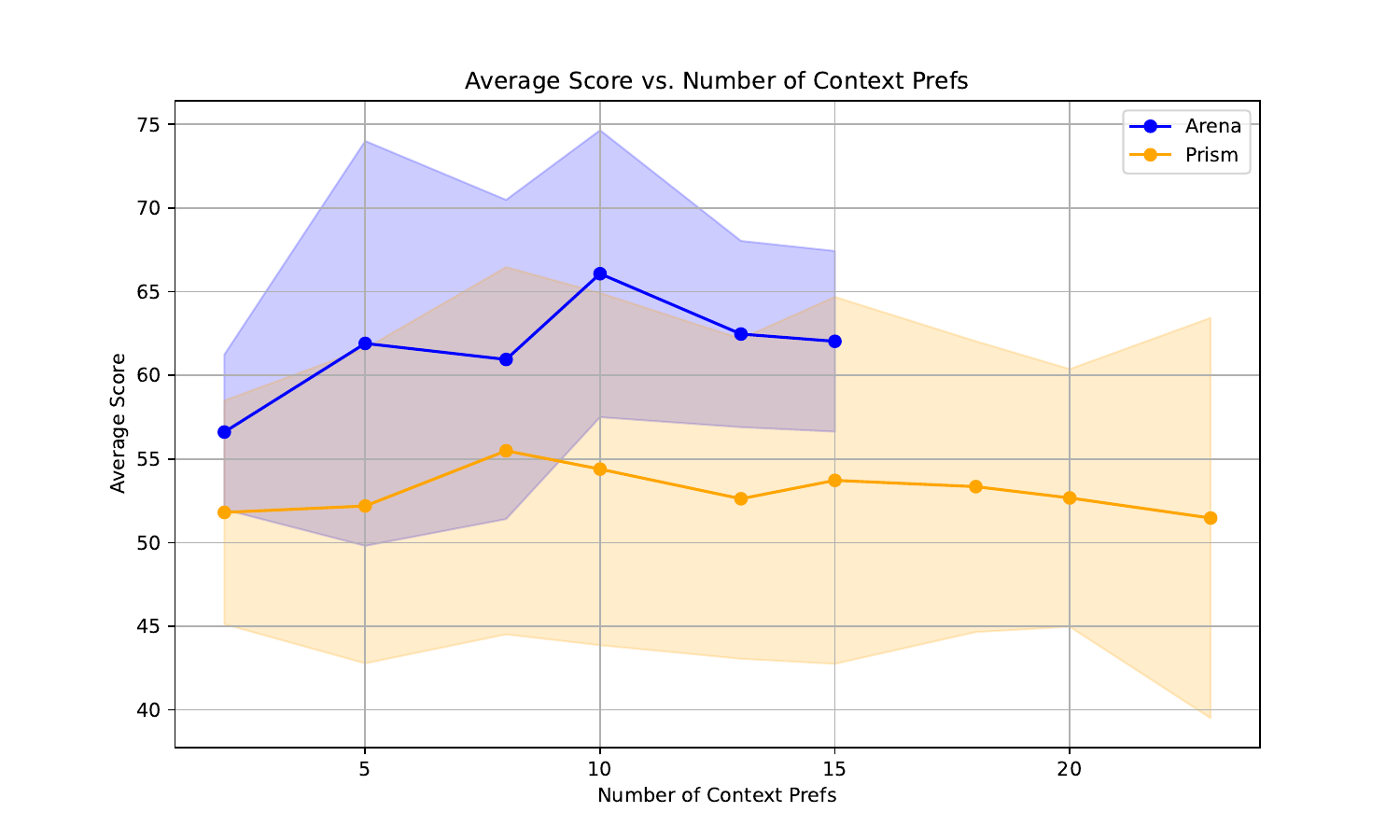}
    \caption{Scaling with prism and arena users with more context.  PRISM does not scale with more context the same way that Chatbot Arena does.  We hypothesize that this is because PRISM users are constrained to 5-6 conversations, so having more interactions just means longer conversations about the same topic, rather than greater diversity of topics.}
    \label{fig:both_datasets_user_scaling}
\end{figure}

\begin{figure}
    \centering
    \includegraphics[width=1.0\linewidth]{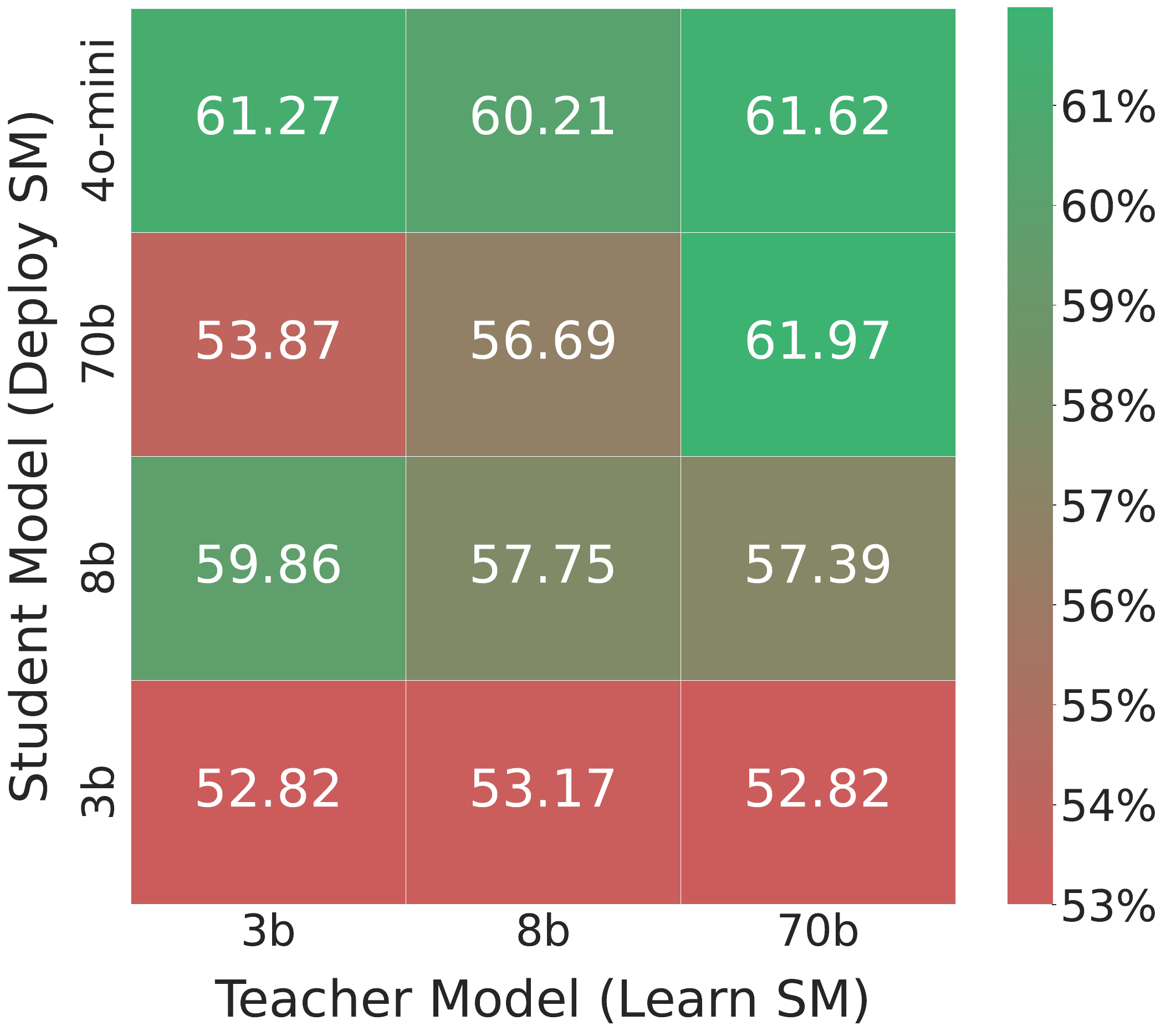}
    \caption{Results of learning a personalized prompts using \texttt{SynthesizeMe} on a teacher model then transferring to a student model for deployment.  Larger models help more for both creating the prompts and executing the personalized reward model.  In the case of chatbot arena the student model appears to have larger impact than the teacher though llama70b is very sensitive to the prompt source. }
    \label{fig:transfer-arena}
\end{figure}

In the main paper we only have so much space to present our results, however having both Chatbot Arena and PRISM means we have twice as many plots per experiment.  In order to streamline the main paper we only include one version of each plot, however we include the extended versions here.  Specifically this section includes a figure for scaling model sizes on chatbot arena (Figure~\ref{fig:arena_scaling}), scaling model sizes on PRISM (Figure~\ref{fig:prism_scaling}), scaling user data on Chatbot Arena and PRISM (Figure~\ref{fig:both_datasets_user_scaling}), and model transfer results on Chatbot Arena (Figure~\ref{fig:transfer-arena}).

\section{Personas versus Demographics}
\label{sec:personas-demographics}

\begin{figure*}[t!] % Force the figure to appear at the top of the page
    \centering
    \includegraphics[width=0.7\textwidth]{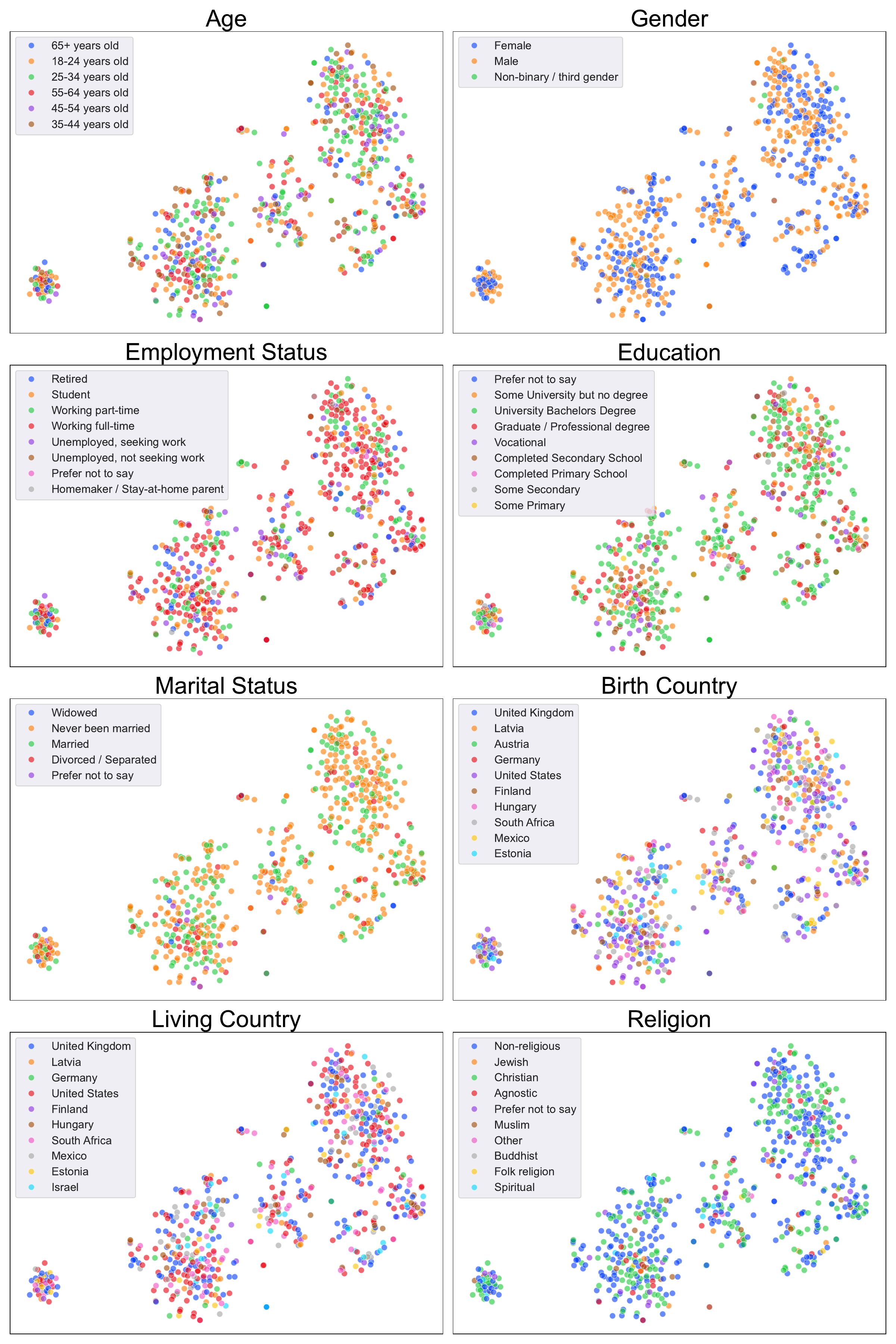}
    \caption{Comparison of demographic categories to t-SNE of \texttt{SynthesizeMe} personas embedding with sBERT}
    \label{fig:persona_tsne_plots}
\end{figure*}

We produce S-BERT embeddings on \texttt{SynthesizeMe} personas produced by Llama-3.1-70B-Instruct and cluster them by demographics using t-SNE dimensionality reduction.  We present these results in Figure~\ref{fig:persona_tsne_plots}.  The most clear insight is that demographics and personas do not clearly cluster, so our personas seem to be capturing different traits than user demographic information.  This is also reflected by the difference in performance between demographic based llm-as-a-judge methods and our context-based methods in Table~\ref{tab:results}.

% Use figure* for two-column layout and control placement

\section{LLM as a Judge and SynthesizeMe Prompts and Programs}
\label{sec:prompts}

\paragraph{DSPy Signatures}
Here, we share the signatures for the DSPy programs that power our LLM as a Judge approaches. Unlike the prompts we share below, these are stable over time and represent the high-level system design of our method.

\begin{figure*}[!t]
  \centering
  \begin{lstlisting}[language=Python]
class LLMAsAJudge(dspy.Signature):
   """Given a conversation and two completions from different models, determine which completion the human judge is more likely to prefer.  Use any provided context to learn about the personal preferences of the judge before making a decision.  If no context is provided it can be useful to speculate about the preferences of the judge.  It's okay to be wrong, let's explore the space of possibilities and hypothesize about what might be true.  Please hypothesize between 1-3 speculations about the judge's preferences or persona when reasoning.  Draw from the context of the conversation and the completions as well as the user written statements to make your decision."""
   conversation: str = dspy.InputField(desc="The conversation context leading up to the completions.")
   first_completion: str = dspy.InputField(desc="The first of the two possible completions to judge between.")
   second_completion: str = dspy.InputField(desc="The second of the two possible completions to judge between.")
   preference: Literal['First', 'Second'] = dspy.OutputField(desc="The completion that the judge is more likely to prefer.  Possible values are 'First' and 'Second'.")
  \end{lstlisting}
  \caption{DSPy Signature for the Default LLM as a Judge Setting used in both initial bootstrapping in SynthesizeMe and benchmarking Default LLM Judge}
  \label{lst:signatures_default_judge}
\end{figure*}

\begin{figure*}[!t]
  \centering
  \begin{lstlisting}[language=Python]
class SynthesizePersona(dspy.Signature):
   """Given a set of user judgements on prior conversations, as well as reasoning for those judgements, concisely build a user persona that can be used to describe the preferences of this person and anything we might know about them."""
   past_judgements:str = dspy.InputField(desc="A set of user judgements on prior conversations alongside reasoning for those judgements.")
   synthesized_persona:str = dspy.OutputField(desc="A synthesized user persona that can be used to inform future judgements.")
  \end{lstlisting}
  \caption{DSPy Signature for Synthesizing Personas from interaction history.  This signature forms the initial prompt before the persona synthesis prompt is optimized.}
  \label{lst:signatures_synthesize}
\end{figure*}

\begin{figure*}[!t]
  \centering
  \begin{lstlisting}[language=Python]
class LLMAsAJudgePersonaInformed(dspy.Signature):
   """Given a conversation and two completions from different models, alongside some prior judgements and a user persona, determine which completion the human judge is more likely to prefer.  Use any provided context as well as the provided persona to speculate about the personal preferences of the judge.  You are a personalized reward model for this user, so think carefully about what this user will like.
   The user you are judging completions for has the FOLLOWING PERSONA: ===
   {persona}
   ===
   
   Now, given the conversation and two completions, decide which completion the user is more likely to prefer.  Remember to consider the user's persona and preferences as you make your decision."""
   conversation:str = dspy.InputField(desc="The conversation context leading up to the completions.")
   first_completion:str = dspy.InputField(desc="The first of the two possible completions to judge between.")
   second_completion:str = dspy.InputField(desc="The second of the two possible completions to judge between.")
   preference:Literal['First', 'Second'] = dspy.OutputField(desc="The completion that the judge is more likely to prefer.  Possible values are 'First' and 'Second'.")
  \end{lstlisting}
  \caption{DSPy Signature for LLM as a Judge with Persona.}
  \label{lst:signatures_persona_judge}
\end{figure*}

\begin{figure*}[!t]
  \centering
  \begin{lstlisting}[language=Python]
class ExtractInsights(dspy.Signature):
    """Given a conversation between a user and an LLM, extract insights from the conversation that can be used to update the user's profile and our understanding of the user's preferences and interests.  If there are no insights, return "no insights found".
    
Insights should be one to two complete sentences in length, and maximally informative about the user."""
    conversation: str = dspy.InputField(desc="The conversation between the user and the LLM.")
    insights: Union[List[str], str] = dspy.OutputField(desc="The insights extracted from the conversation.  If there are no insights, return \"no insights found\".  This should be a list of strings, where each string is an insight.")

class LLMAsAJudgeMemoryInformed(dspy.Signature):
    """Given a conversation and two completions from different models, alongside some prior judgements and a user persona, determine which completion the human judge is more likely to prefer.  Use any provided context as well as the provided persona to speculate about the personal preferences of the judge.  You are a personalized reward model for this user, so think carefully about what this user will like.
    The user you are judging completions for has the FOLLOWING KNOWN FACTS/INSIGHTS: ===
    {memories}
    ===
    
    Now, given the conversation and two completions, decide which completion the user is more likely to prefer.  Remember to consider the user's traits and preferences as you make your decision."""
    conversation:str = dspy.InputField(desc="The conversation context leading up to the completions.")
    first_completion:str = dspy.InputField(desc="The first of the two possible completions to judge between.")
    second_completion:str = dspy.InputField(desc="The second of the two possible completions to judge between.")
    preference:Literal['First', 'Second'] = dspy.OutputField(desc="The completion that the judge is more likely to prefer.  Possible values are 'First' and 'Second'.")
  \end{lstlisting}
  \caption{DSPy Signatures for the memory based personalization experiments.}
  \label{lst:signatures_match}
\end{figure*}

\begin{figure*}[!t]
  \centering
  \begin{lstlisting}[language=Python]
class DetermineMatch(dspy.Signature):
    """Given a self-described preference from a user, and a synthesized persona based on user interactions, determine if the persona is a strong match/fit with this specific user.  If so return True for match, else return False."""

    stated_preferences:str = dspy.InputField(desc="User's stated preferences")
    persona:str = dspy.InputField(desc="User's synthesized persona.  May or may not be a match for this particular user.")
    match:bool = dspy.OutputField(desc="True if the persona matches several of the user's stated preferences, else False")
  \end{lstlisting}
  \caption{DSPy Signature for assessing a match between the stated user preferences and the synthesized persona.}
  \label{lst:signatures_match}
\end{figure*}

\paragraph{Default Prompts}
Here we showcase all the prompts used for the various settings when testing our LLM as a Judge and Synthesize Me prompting approaches.  Note that these prompts serve as a snapshot of a potential prompt to the model, but in reality, the structuring of the adapter is determined at runtime by DSPy \cite{khattab2024dspy}.

\begin{figure*}
\begin{tcolorbox}[colback=blue!5!white, colframe=blue!75!black, title=Default LLM as a Judge Prompt]
\textbf{<<System message:>>} \\
Your input fields are:\\
1. `conversation` (str): The conversation context leading up to the completions.\\
2. `first\_completion` (str): The first of the two possible completions to judge between.\\
3. `second\_completion` (str): The second of the two possible completions to judge between.\\
Your output fields are:\\
1. `reasoning` (str)\\
2. `preference` (Literal['First', 'Second']): The completion that the judge is more likely to prefer.  Possible values are 'First' and 'Second'.\\
All interactions will be structured in the following way, with the appropriate values filled in.\\

[[ \#\# conversation \#\# ]] \\
{conversation} \\

[[ \#\# first\_completion \#\# ]] \\
{first\_completion} \\

[[ \#\# second\_completion \#\# ]] \\
{second\_completion} \\

[[ \#\# reasoning \#\# ]] \\
{reasoning} \\

[[ \#\# preference \#\# ]] \\
{preference}        \# note: the value you produce must exactly match (no extra characters) one of: First; Second \\

[[ \#\# completed \#\# ]] \\
In adhering to this structure, your objective is: \\
        Given a conversation and two completions from different models, determine which completion the human judge is more likely to prefer.  Use any provided context to learn about the personal preferences of the judge before making a decision.  If no context is provided it can be useful to speculate about the preferences of the judge.  It's okay to be wrong, let's explore the space of possibilities and hypothesize about what might be true.  Please hypothesize between 1-3 speculations about the judge's preferences or persona when reasoning.  Draw from the context of the conversation and the completions as well as the user written statements to make your decision. \\

\textbf{<<User message:>>}

[[ \#\# conversation \#\# ]] \\
\{conversation\} \\

[[ \#\# first\_completion \#\# ]] \\
\{first\_completion\} \\

[[ \#\# second\_completion \#\# ]] \\
\{second\_completion\} \\

Respond with the corresponding output fields, starting with the field '[[ \#\# reasoning \#\# ]]', then '[[ \#\# preference \#\# ]]' (must be formatted as a valid Python Literal['First', 'Second']), and then ending with the marker for '[[ \#\# completed \#\# ]]'.
\end{tcolorbox}
\end{figure*}

\begin{figure*}
\begin{tcolorbox}[colback=blue!5!white, colframe=blue!75!black, title=Demographic LLM as a Judge Prompt]
Given the user profile provided below, select the response from AI assistant A or B that the user would most likely prefer.
Declare your choice by using the format: "[[A]]" if you believe assistant A’s response is more suitable, or "[[B]]" if assistant B’s response is better suited. Additionally, assess your confidence in this decision by assigning a certainty level from 1 to 100. Use the following guidelines to assign the certainty level: \\

1–20 (Uncertain): The user profile provides insufficient or minimal evidence. The decision is largely based on weak or indirect hints.\\
21–40 (Moderately Confident): There is noticeable evidence supporting a preference, though it is not comprehensive, and other interpretations are possible.\\
41–60 (Quite Confident): You find clear and convincing evidence that supports your prediction, though it is not entirely decisive.\\
61–80 (Confident): The user profile contains strong evidence that clearly supports your prediction, with very little ambiguity.\\
81–100 (Highly Confident): The user profile provides direct and explicit evidence that decisively supports your prediction.\\
Ensure you enclose your chosen certainty level in double brackets, like so: [[X]].\\

[User Profile] \\
Age: \{age\} Gender: \{gender\} Employment Status: \{employment\_status\} Education: \{education\} Marital Status: \{marital\_status\} Birth Country: \{birth\_country\} Living Country: \{living\_country\} Religion: \{religion\} \\

[User Question] \\
\{question\} \\

[The Start of Assistant A's Answer] \\
\{asst A answer\} 

[The End of Assistant A's Answer] \\

[The Start of Assistant B's Answer] \\
\{asst B answer\} 

[The End of Assistant B's Answer] \\

[Answer] [[ \\
\end{tcolorbox}
\end{figure*}

\begin{figure*}
\begin{tcolorbox}[colback=blue!5!white, colframe=blue!75!black, title=Default Persona Synthesis Prompt]
\textbf{<<System message>>}: \\

Your input fields are: \\
1. 'past\_judgements' (str): A set of user judgements on prior conversations alongside reasoning for those judgements. \\
Your output fields are: \\
1. 'reasoning' (str) \\
2. 'synthesized\_persona' (str): A synthesized user persona that can be used to inform future judgements. \\
All interactions will be structured in the following way, with the appropriate values filled in. \\

[[ \#\# past\_judgements \#\# ]] \\
\{past\_judgements\} \\

[[ \#\# reasoning \#\# ]] \\
\{reasoning\} \\

[[ \#\# synthesized\_persona \#\# ]] \\
\{synthesized\_persona\} \\

[[ \#\# completed \#\# ]] \\
In adhering to this structure, your objective is: \\
        Given a set of user judgements on prior conversations, as well as reasoning for those judgements, concisely build a user persona that can be used to describe the preferences of this person and anything we might know about them. \\

\textbf{<<User message>>}: \\

[[ \#\# past\_judgements \#\# ]] \\
\{past\_judgments\} \\

Respond with the corresponding output fields, starting with the field `[[ \#\# reasoning \#\# ]]`, then `[[ \#\# synthesized\_persona \#\# ]]`, and then ending with the marker for `[[ \#\# completed \#\# ]]`. \\
\end{tcolorbox}
\end{figure*}

\paragraph{Optimized Prompts}
 \label{sec:prompt-optimization}

We also include the DSPy optimized prompts from the MIPROv2 \cite{opsahl-ong-etal-2024-optimizing}. These prompts serve as the Optimized $\Theta$ discussed in $\S$\ref{sec:synthesize-me}.

 We use our training $\mathcal{U}_\text{train}$ and validation $\mathcal{U}_\text{val}$ users as input data for the MIPROv2 \cite{opsahl-ong-etal-2024-optimizing} optimizer and evaluate its performance using improvement on the validation user's target preferences $\mathcal{D}_u^{tgt}$ over default LLM-as-a-judge accuracy.  
 
 Put simply, the MIPROv2 optimizer runs a similar procedure to our rejection sampling for reasoning from steps 1 and 3 (See Figure~\ref{fig:icrm-method}, however it does the same form of rejection sampling on many prompt candidates, finding cases where the prompt succeeds at improving performance.  We measure the performance of a generated persona by looking at how much it improves the performance of a \texttt{SynthesizeMe} prompted LLM Judge over a non-personalized baseline for our validation user. To measure the performance of a novel prompt $\Theta$ we test how this improves persona generation across all validation users $\mathcal{U}_{val}$.   The optimizer uses personas generated on the train users as few-shot demonstrations and tests whether these personas help generate more expressive personas on the validation set.  Note that we are not using DSPy optimization to rewrite the persona string itself, but rather optimizing the persona generation prompt, $\Theta$, to synthesize future personas.  In this way, we only need to incur the costly optimization once, but we benefit from higher quality personas on all future runs. We generate 20 candidate prompts and 20 candidate fewshot-demonstration sets and run MIPROv2 for 30 trials.  We ran the optimizer with max\_bootstrapped\_demos=10, max\_labeled\_demos=6, view\_data\_batch\_size=1, and minibatch\_full\_eval\_steps=3.  Below, we include the optimized prompts for Llama 3B, Llama 8B, and Llama 70B.

\begin{figure*}
\begin{tcolorbox}[colback=blue!5!white, colframe=blue!75!black, title=Llama 3B Optimized Persona Synthesis Prompt (Simplified)]
Propose an instruction to prompt a Language Model to generate a synthesized user persona based on a set of user judgements and reasoning, taking into account the values of appreciation, contentment, and openness to new experiences. \\

[[Past\_Judgements]] \\
\{Example User 1 History\} ... \\

[[Reasoning]] \\
The user appears to value respect, openness, and individuality, as seen in their responses to conversations about trans issues and relationships. They prioritize personal autonomy and the importance of respecting one's own feelings and needs. The user also seems to be open to learning and considering different perspectives, as evidenced by their willingness to explore topics they may not be familiar with. \\

[[Synthesized\_Persona]] \\
This user is likely someone who values independence, self-awareness, and respect for others' differences. They may be introverted or prefer to focus on their own interests and goals, but are still open to engaging with others and learning from their experiences. They prioritize their own emotional well-being and may be hesitant to commit to relationships or social interactions that don't feel authentic or fulfilling to them. \\

[[Past\_Judgements]] \\
\{Example User 2 History\} ... \\

[[Reasoning]] \\
The judge values nuance and balance in discussions, prefers clear and respectful communication, and is more likely to engage with complex and thought-provoking topics. They also seem to be empathetic and understanding, acknowledging the complexity of issues and the need for personal belief and value judgments. \\

[[Synthesized\_Persona]] \\
A thoughtful and empathetic individual who values nuance and balance in discussions. They are likely to be someone who is familiar with the challenges of navigating complex social relationships and is more likely to appreciate a personalized and adaptive approach to problem-solving. This person may be someone who is skeptical of generic or one-size-fits-all solutions and is more likely to engage with complex and thought-provoking topics. \\

[[Past\_Judgements]] \\
\{Example User 3 History\} ... \\

[[Reasoning]] \\
\{Example User 3 Reasoning\} \\

[[Synthesized\_Persona]] \\
\{Example User 3 Synthesized Persona\} \\

... For a total of 6 Example Synthesized Personas ...
\end{tcolorbox}
\end{figure*}

\begin{figure*}
\begin{tcolorbox}[colback=blue!5!white, colframe=blue!75!black, title=Llama 8B Optimized Persona Synthesis Prompt]
Propose an instruction that will be used to prompt a Language Model to generate a user persona based on a set of past judgements and reasoning. The instruction should include a high-stakes scenario in which the LM must solve the task. \\

Imagine you are a detective tasked with solving a mysterious case involving a series of eerie and surreal events. You have been provided with a set of past judgements and reasoning from a previous investigation, and you must use this information to build a user persona that will help you crack the case. The persona should be based on the individual's preferences, values, and behaviors, and should provide insight into their motivations and actions. The goal is to create a comprehensive and accurate user persona that will aid in the investigation and ultimately lead to the resolution of the case. Please generate a user persona based on the provided information, and be prepared to defend your reasoning and conclusions. \\

\{Example User 1 History\} ... \\

[[Reasoning]] \\
Based on the provided conversations, the user seems to prefer more detailed and comprehensive responses. They tend to engage with longer answers that provide multiple suggestions or explanations, rather than brief ones. This is evident in their preference for the second completion in conversations about quieting the mind during meditation and the philosophical question of responsibility for immoral actions. Additionally, the user's preference for the first completion in the conversation about music recommendations suggests that they value a more interactive and personalized approach. Overall, the user appears to appreciate thoughtful and detailed responses that cater to their specific needs and interests. \\

[[Synthesized\_Persona]] \\
Based on the user's preferences and interactions, a synthesized persona can be created as follows:\textbackslash n\textbackslash n Name: Reflective\textbackslash n\textbackslash n Age: Late 20s to early 30s\textbackslash n\textbackslash nPersonality: Reflective is a thoughtful and introspective individual who values depth and nuance in their conversations. They tend to engage with complex ideas and appreciate detailed explanations. Reflective is patient and kind, often taking the time to consider multiple perspectives before forming an opinion.\textbackslash n \textbackslash nInterests: Meditation, philosophy, music, and self-improvement\textbackslash n\textbackslash n Goals: Reflective aims to cultivate a deeper understanding of themselves and the world around them. They strive to develop a more compassionate and empathetic approach to life, and to find balance and harmony in their thoughts and actions.\textbackslash n\textbackslash n Values: Thoughtfulness, nuance, patience, kindness, and self-awareness \\

\{Example User 2 History\} ... \\

[[Reasoning]]
\{Example User 2 Reasoning\} \\

[[Synthesized\_Persona]] \\
Meet "Alex," a 30-year-old individual who is deeply invested in their spiritual growth and critical thinking. They value independence and autonomy in their intellectual pursuits, often preferring to explore complex topics on their own rather than relying on external guidance. Alex is a curious and introspective person who appreciates the importance of considering multiple perspectives and historical context when interpreting religious texts. They are also passionate about learning and retaining information, and believe that language barriers can be a significant obstacle to forming meaningful connections with people from diverse backgrounds. Alex is likely to be drawn to careers or activities that promote critical thinking, cultural exchange, and language learning. \\

... For a total of 6 Example Synthesized Personas ...
\end{tcolorbox}
\end{figure*}

\begin{figure*}
\begin{tcolorbox}[colback=blue!5!white, colframe=blue!75!black, title=Llama 70B Optimized Persona Synthesis Prompt]
Given a collection of conversations between a user and a model, where the user has expressed their preferences and emotions through their interactions, and considering the reasoning behind their judgements on these conversations, create a comprehensive and empathetic user persona that captures their unique characteristics, values, and interests. This persona should reflect the user's need for personalized and interactive responses, their appreciation for contemporary and modern approaches, and their desire for supportive and informative guidance on topics such as art, social justice, and environmental issues. The persona should also acknowledge the user's struggles with pessimism and hopelessness, and highlight their motivation to create positive change. By analyzing the user's past conversations and judgements, generate a detailed and nuanced persona that can be used to inform future interactions and provide tailored support and recommendations. \\

\{Example User 1 History\} ... \\

[[Reasoning]] 
The user's preferences and judgements on prior conversations can be analyzed to synthesize a persona. In the conversations provided, the user showed a preference for concise and direct answers, as seen in their preference for the first completion in multiple conversations. They also demonstrated an interest in learning about the night sky, traveling in Europe, and experiencing different cultures. The user's preference for the first completion in the conversation about traveling from London to the Netherlands suggests that they value efficiency and convenience. Additionally, their preference for the first completion in the conversation about the best way to travel around Europe indicates that they prioritize ease and affordability. \\

[[Synthesized\_Persona]] 
The synthesized persona is a curious and practical individual who values efficiency, convenience, and affordability. They are interested in learning about the world around them, including the night sky and different cultures. When traveling, they prioritize ease and affordability, preferring to take trains or use convenient transportation options. They also appreciate direct and concise answers, suggesting that they are busy and value their time. This persona can be used to inform future judgements and provide more tailored responses to their queries. \\

\{Example User 2 History\} ... \\

[[Reasoning]] 
The user's preferences can be inferred from their judgements on prior conversations. In the first two conversations about political apathy, the user preferred the first completion, which provided a more concise and direct answer. However, in the conversations about managing work and family life, the user preferred the second completion, which provided a more practical and step-by-step approach. This suggests that the user values clarity and concision in their responses, but also appreciates practical advice and solutions. Additionally, the user's preferences may vary depending on the topic and context of the conversation. \\

[[Synthesized\_Persona]] 
Based on the user's preferences, a synthesized persona can be created. This persona values clarity, concision, and practicality in their interactions. They are likely busy professionals or individuals with multiple responsibilities, who need efficient and effective solutions to manage their work and family life. They are also interested in social and political issues, but may not have the time or energy to engage deeply with complex or abstract concepts. They prefer direct and straightforward answers, but also appreciate nuanced and thoughtful responses that take into account multiple perspectives. This persona is likely motivated by a desire to balance their personal and professional responsibilities, while also staying informed and engaged with the world around them. \\

... For a total of 5 Example Synthesized Personas ...

\end{tcolorbox}
\end{figure*}

\section{Learned Personas through SynthesizeMe}

Here we provide the personas for a randomly sampled user, user1118, from the PRISM test set.  We provide personas from Llama 3.2 3B, Llama 3.1 8B, Llama 3.3 70B, Gemini-2.0-Flash, Gemini-2.5-Flash, Gemini-2.5-Pro, GPT-4o-mini, Qwen3-8B, Qwen3-30B-3BA, and Qwen3-32B for comparison.

\begin{figure*}
\begin{tcolorbox}[colback=blue!5!white, colframe=blue!75!black, title=Llama 3.2 3B -- User1118 Persona]
The synthesized persona based on the user's judgements and preferences is that of an individual who values helpful and informative responses. This person is likely to be open to suggestions and ideas, and appreciates clear and concise communication. They seem to prioritize practical and useful information, as seen in their interest in recipes and suggestions for making the most of a kale salad. The persona is also likely to be enthusiastic and engaged, as seen in their positive response to the second completion's suggestion of using alternative crunchy ingredients. While they may occasionally prefer more direct or affirmative responses, their overall preference leans towards comprehensive and considerate discussions that reflect the complexity of real-world issues.
\end{tcolorbox}
\end{figure*}

\begin{figure*}
\begin{tcolorbox}[colback=blue!5!white, colframe=blue!75!black, title=Llama 3.1 8B -- User1118 Persona]
The synthesized persona based on the user's judgements and preferences is that of an individual who values clear, concise, and helpful information. This person likely appreciates responses that are direct, enthusiastic, and provide the information they need. They seem to prioritize practicality and usefulness over elaborate or overly detailed responses. The persona is likely someone who is interested in learning and exploring new ideas, but values efficiency and effectiveness in their interactions. While they may occasionally prefer more detailed or elaborate responses, their overall preference leans towards clear and helpful discussions that provide the information they need.
\end{tcolorbox}
\end{figure*}

\begin{figure*}
\begin{tcolorbox}[colback=blue!5!white, colframe=blue!75!black, title=Llama 3.3 70B -- User1118 Persona]
The synthesized persona based on the user's preferences is someone who values helpfulness, informativeness, and engagement in conversations. This person likely appreciates detailed and practical information, such as recipes and instructions, and enjoys exploring new ideas and suggestions. They seem to respond positively to friendly and enthusiastic language, and may be more likely to engage in conversations that are welcoming and encouraging. The persona is likely someone who is interested in cooking, sports, and travel, and values conversations that are informative, yet casual and approachable. They may also appreciate models that can provide creative suggestions and substitutions, and are open to trying new things.
\end{tcolorbox}
\end{figure*}

\begin{figure*}
\begin{tcolorbox}[colback=blue!5!white, colframe=blue!75!black, title=Gemini-2.0-Flash -- User1118 Persona]
The synthesized persona is someone who appreciates direct, specific, and informative responses. They are likely engaging and enjoy a conversational tone. They value helpfulness and encouragement, and respond positively to enthusiasm and a welcoming attitude. This person is likely curious and open to new ideas, and appreciates when others are willing to share their knowledge and experiences.
\end{tcolorbox}
\end{figure*}

\begin{figure*}
\begin{tcolorbox}[colback=blue!5!white, colframe=blue!75!black, title=Gemini-2.5-Flash -- User1118 Persona]
The user is a pragmatic and direct individual who values clear, concise, and immediately actionable information. They prefer responses that get straight to the point, providing specific details or a nuanced, qualified stance rather than generalities or AI disclaimers. They appreciate efficiency and relevance, disliking extraneous information or overly verbose explanations. This user seeks practical utility from the model and responds well to a helpful, encouraging, and welcoming tone, especially when it accompanies concrete advice or information.
\end{tcolorbox}
\end{figure*}

\begin{figure*}
\begin{tcolorbox}[colback=blue!5!white, colframe=blue!75!black, title=Gemini-2.5-Pro -- User1118 Persona]
This user seeks personalized and interactive responses that are both supportive and directly informative. They appreciate contemporary approaches, valuing clarity, efficiency, and actionable information, as demonstrated by their preference for immediate recipe details over conversational preamble. When discussing cultural interests, such as the "Welcome to Wrexham" series, they respond well to an enthusiastic, welcoming, and slightly more personal tone, indicating a desire for engaging and relatable interactions. However, this appreciation for warmth is balanced by a strong need for relevance and conciseness; off-topic or rambling content is not well-received. \\

On more complex subjects, such as the societal impact of organized religion (which touches on social values and justice), the user desires nuanced and balanced guidance. They are receptive to responses that thoughtfully consider different perspectives and offer conditional viewpoints, suggesting an interest in understanding multifaceted issues rather than receiving overly simplistic or purely detached statements. This indicates an appreciation for informative content that can support their own considerations and potentially inform their views on creating positive change. They are looking for an AI that can engage meaningfully, providing clear, useful, and thoughtfully considered information across a range of topics, including art, social issues, and practical guidance.

\end{tcolorbox}
\end{figure*}

\begin{figure*}
\begin{tcolorbox}[colback=blue!5!white, colframe=blue!75!black, title=GPT4o-mini -- User1118 Persona]
The synthesized persona is someone who values nuanced and balanced discussions, particularly on complex topics such as organized religion and social issues. They appreciate responses that acknowledge multiple perspectives and provide a comprehensive view. This persona is also likely to enjoy engaging conversations, as evidenced by their preference for responses that invite further interaction, especially in casual contexts like cooking or entertainment. They prefer clear and direct answers that convey enthusiasm and positivity, while also being open to exploring different viewpoints. This persona is curious, thoughtful, and values respectful dialogue, making them receptive to informative and engaging exchanges.
\end{tcolorbox}
\end{figure*}

\begin{figure*}
\begin{tcolorbox}[colback=blue!5!white, colframe=blue!75!black, title=Qwen3-8B -- User1118 Persona]
The synthesized persona is a detail-oriented, practical individual who values actionable advice and personalized solutions. They seek balanced, nuanced information on complex topics, appreciating responses that acknowledge multiple perspectives while avoiding bias. They enjoy cooking and are open to experimenting with recipes, valuing creativity and adaptability in culinary contexts. In broader discussions, they prioritize objectivity, respect for diverse viewpoints, and clear communication. They are likely to engage thoughtfully with topics like social issues, relationships, and personal growth, preferring interactions that are both informative and supportive. Their preferences suggest a desire for practicality, empathy, and a willingness to explore nuanced ideas without oversimplification.

\end{tcolorbox}
\end{figure*}

\begin{figure*}
\begin{tcolorbox}[colback=blue!5!white, colframe=blue!75!black, title=Qwen3-30B-3BA -- User1118 Persona]
The synthesized persona is an inquisitive, balanced individual who values nuanced, well-reasoned responses. They seek practical, specific information (e.g., detailed recipes, contextual analysis of media) and appreciate discussions that acknowledge complexity rather than oversimplifying issues. They are likely critical of irrelevant or overly optimistic content and prefer answers that reflect a thoughtful understanding of multifaceted topics. Their interests span cultural, social, and practical domains, and they engage with information in a way that prioritizes depth, relevance, and fairness.
\end{tcolorbox}
\end{figure*}

\begin{figure*}
\begin{tcolorbox}[colback=blue!5!white, colframe=blue!75!black, title=Qwen3-32B -- User1118 Persona]
The synthesized persona is a **\textbf{curious, practical, and socially engaged individual}** who values:  \\
1. **\textbf{Nuanced, balanced perspectives}** on complex topics (e.g., religion, relationships).  \\
2. **\textbf{Specific, actionable details}** in practical contexts (e.g., recipes, travel recommendations).  \\
3. **\textbf{Relevance and coherence}** in responses, avoiding tangents or vague generalizations.  \\
4. **\textbf{Empathy and community focus}**, showing interest in shared experiences (e.g., football culture, fan passion).  \\
5. **\textbf{Efficiency in communication}**, preferring concise yet thorough explanations over fluff.  \\

This persona is likely a **\textbf{lifelong learner}** who enjoys exploring diverse topics but prioritizes clarity and utility in information. They are open to multiple viewpoints but expect responses to acknowledge complexity without oversimplification.
\end{tcolorbox}
\end{figure*}

% \clearpage

% \section{Open Feedback}
% \FloatBarrier % Prevent figures from appearing above this point

% \begin{figure}[H] % Force the figure to appear here
%     \centering
%     \includegraphics[width=0.98\linewidth]{img/models.png}
%     \caption{Models used in generating completions in our Open Feedback data collection.}
%     \label{fig:models_open_feedback}
% \end{figure}

% \begin{figure*}
%     \centering
%     \includegraphics[width=0.8\textwidth]{img/chat-ui-conversation.pdf}
%     \caption{Conversation in the user interface for Open Feedback}
%     \label{fig:chat-ui-conversation}
% \end{figure*}

% \begin{figure*}
%     \centering
%     \includegraphics[width=0.8\textwidth]{img/chat-ui-completions.pdf}
%     \caption{Pairwise completions in the user interface.  Two distinct completions are presented for selection whenever the user prompts the model.}
%     \label{fig:chat-ui-completions}
% \end{figure*}

% \begin{figure*}
%     \centering
%     \includegraphics[width=0.6\textwidth]{img/chat-ui-feedback.pdf}
%     \caption{Feedback interface for gathering user feedback.  Users can provide feedback on any of the pairwise completions.}
%     \label{fig:chat-ui-feedback}
% \end{figure*}

% \begin{figure*}
%     \centering
%     \includegraphics[width=0.8\textwidth]{img/chat-ui-edit.pdf}
%     \caption{Edit screen for allowing the user to modify model outputs.}
%     \label{fig:chat-ui-edit}
% \end{figure*}

\end{document}